\RequirePackage{fix-cm}

\documentclass{svjour3}                     
\smartqed  

\usepackage{natbib}
\setcitestyle{authoryear,open={(},close={)}}
\usepackage{subcaption}
\usepackage[utf8]{inputenc} 
\usepackage[T1]{fontenc}    
\usepackage{hyperref}       
\usepackage{url}            
\usepackage{booktabs}       
\usepackage{amsfonts}       
\usepackage{amsmath}
\usepackage{amssymb}
\usepackage{nicefrac}       
\usepackage{microtype}      
\usepackage{siunitx}
\usepackage{multirow}
\usepackage{breqn}
\usepackage[misc]{ifsym}
\usepackage{dsfont}
\usepackage{extdash}
\usepackage{enumitem}
\usepackage[section]{placeins}

\usepackage{verbatim}
\usepackage[labelfont=bf]{caption}

\newcommand{\corrAuthor}{$^{\textrm{\Letter}}$}

\newtheorem{deff}{Definition}
\usepackage{mathtools}
\usepackage{algorithmic}
\usepackage[ruled,vlined]{algorithm2e}

\newlength\mylen
\newcommand\myinput[1]{%
  \settowidth\mylen{\KwIn{}}%
  \setlength\hangindent{\mylen}%
  \hspace*{\mylen}#1\\}
  
\DeclareMathOperator*{\argmax}{arg\,max}

\journalname{Computational Brain and Behavior}

\DeclareUnicodeCharacter{2212}{-}

\begin{document}
\title{Boosting human decision-making with AI-generated decision aids}

\author{Frederic Becker\textsuperscript{1,*} \and Julian Skirzy\'nski\textsuperscript{1,2,*} \and Bas van Opheusden\textsuperscript{3} \and Falk Lieder\textsuperscript{1}
}

\authorrunning{F. Becker, J. Skirzy\'nski, B. van Opheusden, F. Lieder} 

\institute{\textsuperscript{*} Joint first-authorship\\
            \textsuperscript{1} Max Planck Institute for Intelligent Systems, T\"ubingen, Germany\\
           \textsuperscript{2} University of California, San Diego, CA 92093, USA\\
           \textsuperscript{3} Princeton University, Princeton, NJ 08544, USA \\
            \corrAuthor  \email{frederic.becker@tuebingen.mpg.de} \\
}

\date{Received: date / Accepted: date}

\sisetup{tight-spacing=true}
\maketitle

\begin{abstract}

Human decision-making is plagued by many systematic errors. Many of these errors can be avoided by providing decision aids that guide decision-makers to attend to the important information and integrate it according to a rational decision strategy. 
Designing such decision aids used to be a tedious manual process. Advances in cognitive science might make it possible to automate this process in the future. We recently introduced machine learning methods for discovering optimal strategies for human decision-making automatically and an automatic method for explaining those strategies to people. Decision aids constructed by this method were able to improve human decision-making. However, following the descriptions generated by this method is very tedious.
We hypothesized that this problem can be overcome by conveying the automatically discovered decision strategy as a series of natural language instructions for how to reach a decision.
Experiment~1 showed that people do indeed understand such procedural instructions more easily than the decision aids generated by our previous method. Encouraged by this finding, we developed an algorithm for translating the output of our previous method into procedural instructions. We applied the improved method to automatically generate decision aids for a naturalistic planning task (i.e., planning a road trip) and a naturalistic decision task (i.e., choosing a mortgage). Experiment~2 showed that these automatically generated decision-aids significantly improved people's performance in planning a road trip and choosing a mortgage.
These findings suggest that AI-powered boosting might have potential for improving human decision-making in the real world. 

\textbf{Keywords}: improving human decision-making; boosting; decision aids; far-sightedness; interpretable machine learning; automatic strategy discovery
\end{abstract}

\section{Introduction}

Many researchers working on judgment and decision-making \citep[e.g., ][]{TverskyKahneman74,gilovich2002heuristics} have argued that human decision-making is plagued by many systematic errors known as cognitive biases \citep[but see ][]{gigerenzer1991make,gigerenzer2008cognitive}. In particular, older people often come to regret the short-sighted decisions they made about their health, education, and finances when they were younger \citep{kinnier1989regrets}. Consistent with this observation, numerous experiments on intertemporal choice have consistently found that people's decisions depend primarily on the immediate outcomes of potential choices and underweight their more weighty long-term consequences  \citep{milkman2008harnessing,o2015present}.
Those short-sighted choices can be traced back to the strategies people use to make decisions \citep{reeck2017search}. Therefore, one way to address such problems is to improve people's decision strategies. This is an instance of \textit{boosting} \citep{hertwig2017nudging}. Boosting human decision-making has many benefits over delegating decisions to algorithms. Firstly, there are multiple areas, such as medicine and the judicial system, in which humans are and will continue to be the ultimate decision-makers for ethical reasons. Secondly, people have to make many subjective decisions that depend on their personal values. Helping people make better decisions in these settings is crucial for increasing our society's well-being.

\begin{figure}[b]
    \centering
    \includegraphics[width=0.9\linewidth]{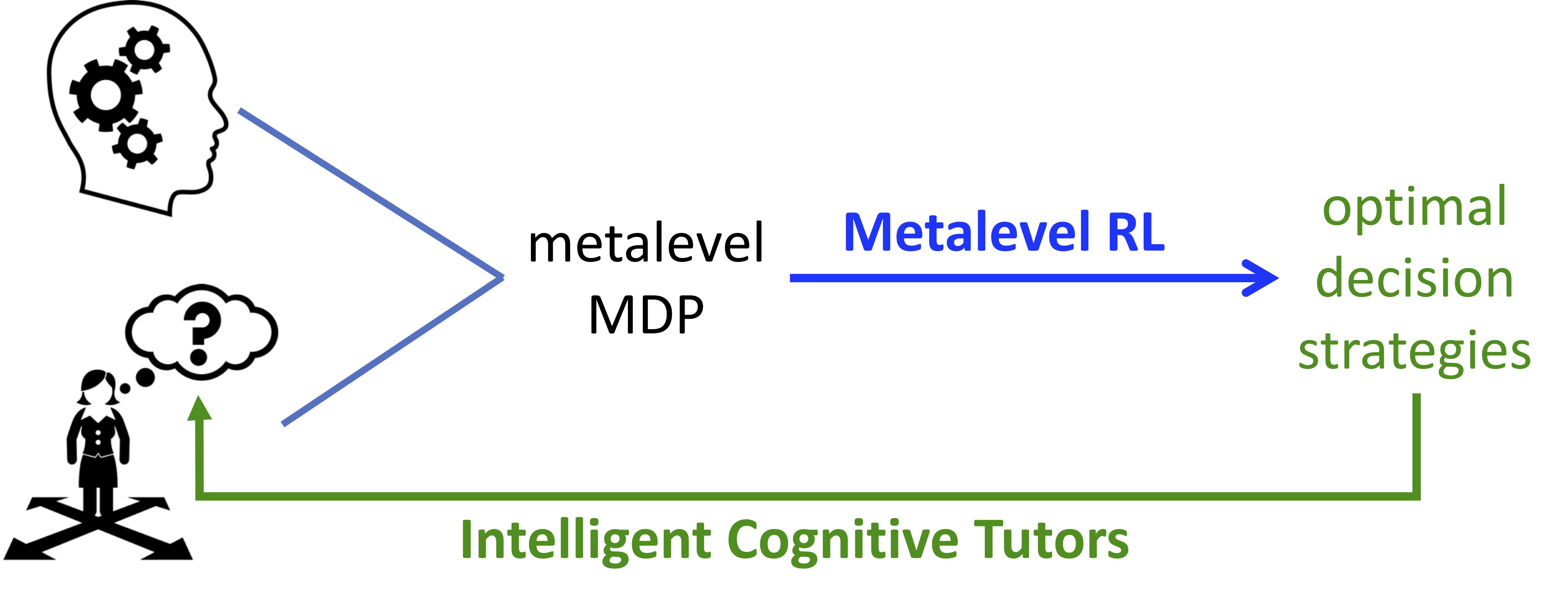}
    \caption{AI-powered boosting relies on discovering optimal decision strategies by modeling decision problems as a metalevel Markov Decision Process and solving them with metalevel reinforcement learning \citep{Griffiths2019}}
    \label{fig:SD}
\end{figure}

Boosting can be implemented by training people or by supporting them while they make a decision. The benefits of training on a simple task rarely transfer to the more complex problems people face in the real world \citep{sala2017does,sala2019near,becker2021encouraging}. One way to side-step this problem is to provide people with decision aids that support them directly in those decisions that are to be improved. Previous research showed that real-world decisions in the domains of finance and medicine can be improved by providing people with decision aids (e.g., a decision tree) that guide them through the application of clever heuristic that direct the decision-maker's attention to the most essential pieces of information \citep{hafenbradl2016applied}. Recent work has developed algorithms for generating and visualizing decision trees automatically  \citep{phillips2017fftrees,rudin2022interpretable}. However, equivalent tools do not yet exist for helping people solve complex planning problems. Designing decision aids for such problems by hand can be very tedious, and coming up with clever heuristics can very difficult. As a first step towards addressing this problem, we recently developed a computational method for automatically designing decision aids for sequential decision problems that require planning  \citep{skirzynski2020automatic}. This AI-powered boosting method leverages Artificial Intelligence (AI) to derive smart decision strategies from a mathematical theory of optimal decision-making with finite time, limited time, and bounded cognitive resources \citep{lieder_griffiths}. The decision aids constructed by this method were able to improve human decision-making \citep{skirzynski2020automatic}. However, using these decision aids is tedious because they do not explicitly specify the decision process directly. Rather, they specify a process for determining whether the next step the decision-maker is considering to take is consistent with the recommended heuristic. The goal of this article is to develop an improved method that can generate decision aids that directly describe the automatically discovered heuristic in natural language.

\begin{figure}[h!]
    \centering
    \includegraphics[width=0.8\linewidth]{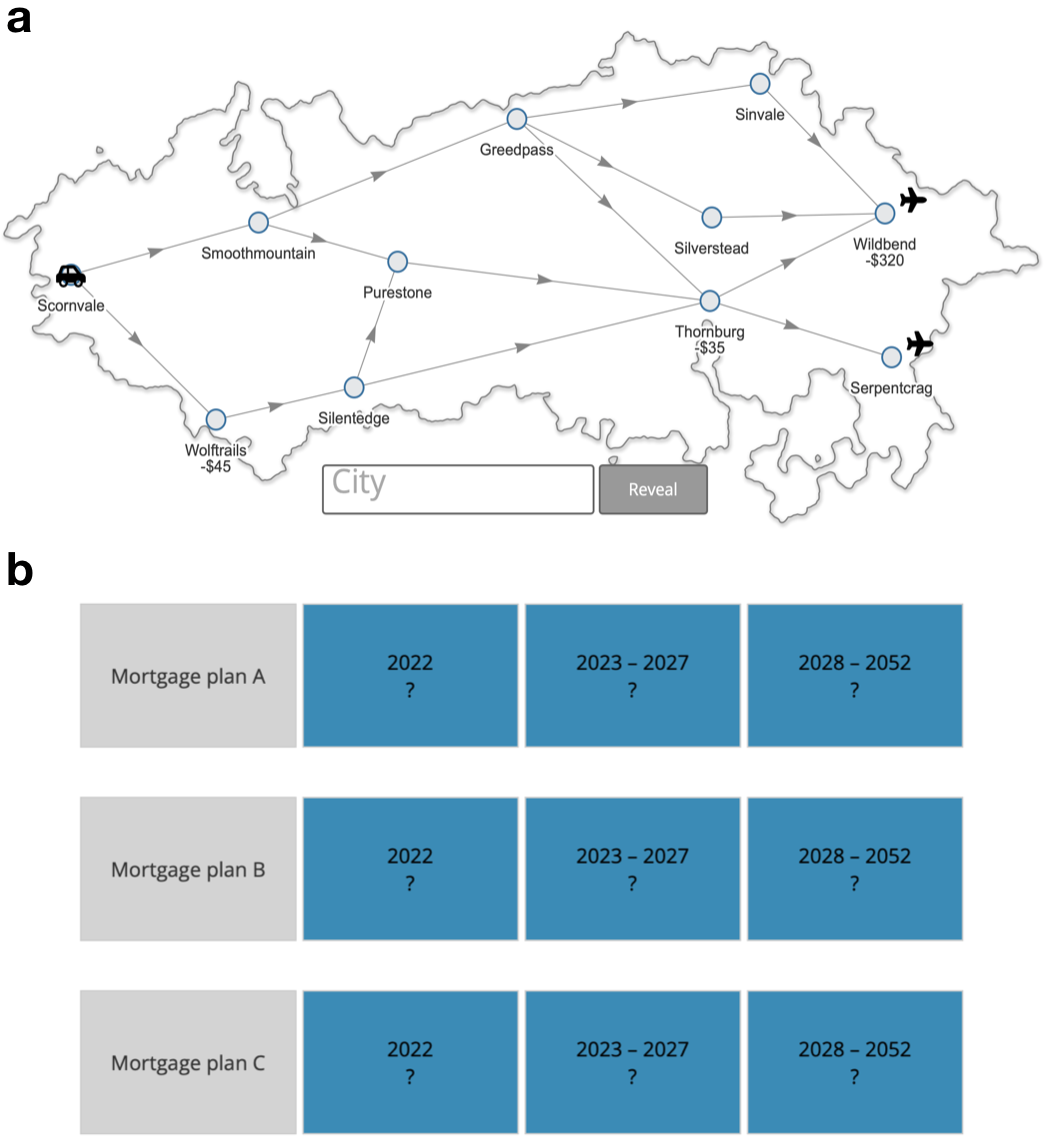}
    \caption{Naturalistic decision-tasks used in Experiment~2. \textbf{a}: In the Road Trip task, the objective is to efficiently find an inexpensive route to a city with an airport. Participants can use a search engine to look up how costly it is to spend the night in different cities. \textbf{b}: In the Mortgage task the goal is to choose the most affordable mortgage by considering interest rates for different time horizons. Participants can learn about the interest rates by clicking on appropriate boxes}
    \label{fig:envs}
\end{figure}

We hypothesized that step-by-step instructions for how to reach a decision would be significantly easier for people to follow than the decision aids generated by our previous method \citep{skirzynski2020automatic}. After confirming that such \textit{procedural instructions} are more interpretable and easier to follow than the previous version of our decision aids, we developed a new algorithm for transforming the output of the previous method \citep{skirzynski2020automatic} into procedural instructions. The extended method automatically generates step-by-step, natural language instructions for how to reach a decision. These instructions are generated from a pair of two inputs: i) a model of the general structure of a particular (sequential) decision problem, and ii) a dictionary of what the relevant components of the general problem are called in the concrete application. This approach is very general and can be applied to different kinds of decision problems. In particular, the two algorithms we utilized to set up our approach give rise to a new policy-agnostic method for interpreting reinforcement learning policies, which is an important part of the problem known as explainable reinforcement learning \citep{puiutta2020explainable, dazeley2021explainable}. As a proof-of-concept, we demonstrate that our method can be used to make more far-sighted choices in two naturalistic decision-tasks: planning a road trip and choosing a mortgage (see Fig.~\ref{fig:envs}a and Fig.~\ref{fig:envs}b).
We found that people can understand and follow the automatically generated procedural instructions in both tasks and consequently made better decisions. These findings suggest that AI-powered boosting can be very effective at improving human decision-making in naturalistic tasks. 

The article is structured as follows. In the next section we provide further information on approaches striving to improve human decision making and discuss the connections between our method and other approaches to decision support. In the third section, we show that procedural descriptions of planning strategies are better suited to improve human decision making than the descriptions generated by the previous method \citep{skirzynski2020automatic}. In the following section, we present a new method for automatically generating procedural descriptions of planning strategies. In Section 5, we test the extent to which decision aids obtained with our method improve human decision making in more naturalistic decision problems. Lastly, we summarize and discuss our findings in Section 6.

\section{Background}

\subsection{Reinforcement learning}
To understand the underlying principles and mechanisms of AI-powered boosting, we provide a handful of definitions for mathematical constructs that are used by this approach. Those constructs regard the theory of reinforcement learning that enables to compute solutions to sequential decision problems that are provably optimal. Moreover, these methods play an important role in our approach to discovering optimal cognitive strategies for human decision-making.

The first definition presents a structure that is a general model of decision-making problems called \emph{Markov Decision Process} (see Definition~\ref{deff:MDP}). Representing a decision problem in terms of a Markov Decision Process becomes possible by specifying what states exist in this problem, what actions might be taken, how valuable (rewarding) each of the actions is in the states, and how the actions change the states form one to another.
\begin{deff}[Markov Decision Process]\label{deff:MDP}
A Markov decision process (MDP) is a tuple $(\mathcal{S},\mathcal{A},\mathcal{T}, \mathcal{R}, \gamma$) where $\mathcal{S}$ is a set of states; $\mathcal{A}$ is a set of actions; $\mathcal{T}(s,a,s') = \mathbb{P}(s_{t+1}=s'\mid s_t = s, a_t = a)$ for $s\neq s'\in\mathcal{S}, a\in\mathcal{A}$ is a state transition function; $\gamma\in (0,1)$ is a discount factor; $\mathcal{R}:\mathcal{S}\to\mathbb{R}$ is a reward function.
\end{deff}

Having modeled a decision problem as an MDP, we may further define a mathematical counterpart of a decision strategy, namely a \emph{policy} (see Definition~\ref{deff:policy}). 
\begin{deff}[Policy]\label{deff:policy}
A deterministic policy $\pi$ is a function $\pi: \mathcal{S}\rightarrow\mathcal{A}$ that specifies actions to take in each of the states in the MDP and a non-deterministic policy $\pi$ is a function $\pi: \mathcal{S}\rightarrow Prob(\mathcal{A})$ that defines a probability distribution over the actions for the states in the MDP. 
\end{deff}

Then, \emph{expected reward} (see Definition~\ref{deff:reward}) serves to quantify the usefulness of policies (i.e. to quantify how valuable or rewarding they are).
\begin{deff}[Expected reward]\label{deff:reward}
The reward $r_t$ represents the quality of performing action $a_t$ in state $s_t$. The cumulative return of a policy is a sum of its discounted rewards obtained in each step of interacting with the MDP, i.e. $G_t^{\pi}=\sum\limits_{i=t}^{\infty}\gamma^t r_t$ for $\gamma\in[0,1]$. The expected reward $J(\pi)$ of policy $\pi$ is equal to $J(\pi)=\mathbb{E}(G_0^{\pi})$.
\end{deff}

Finally, equipped with these formalizations of decision strategies and their quality in the problem modeled by the MDP, we need an approach for finding strategies whose quality is high. A family of methods which solve that problem and find policies with the highest reward is called \emph{reinforcement learning} (see Definition~\ref{deff:rl}). 
\begin{deff}[Reinforcement learning]\label{deff:rl}
Reinforcement learning (RL) is a class of methods that perform iterations over trials and evaluation on a given MDP to find the optimal policy $\pi^*$ which maximizes the expected reward \citep{Sutton2018}.
\end{deff}

\subsection{Modeling planning as information acquisition}\label{sec:mouselab_mdp}

\begin{figure}
\centering
\includegraphics[width=0.55\linewidth]{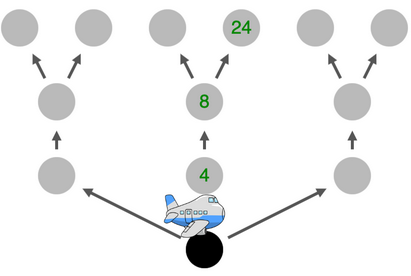}
\caption{Mouselab-MDP with increasing variance. Nodes in the Mouselab-MDP represent short-, mid-, and long-term consequences of potential actions where the black node is the starting position, and connections in the Mouselab-MDP symbolize ``possible later consequence'' relation. Information on the consequences is represented numerically as rewards, and can be acquired by clicking on the nodes. Rewards are drawn from a normal distribution whose variance increases with the distance of the node from the starting node.  Thus, rewards that are farthest from the starting position (long-term consequences) show the highest variance. As gathering information is costly, the agent's goal in the Mouselab-MDP is to make as few clicks as possible to uncover the best possible course of actions with the highest total reward}
\label{fig:mouselab}
\end{figure}

In our quest to improve human planning, we used a theoretical framework developed to study and improve human planning, namely the resource-rational theory of optimal human planning \citep{Callaway2022Rational,callaway2018resource}. According to this theory, planning can be understood as a series of information gathering operations that gradually refine a person's estimates of the short- and long-term consequences of alternative courses of action. This idea gave rise to an empirical paradigm (i.e., the Mouselab-MDP paradigm) that operationalizes human planning by the series of clicks made to collect information about the rewards hidden at different locations of the environment of a path planning task (see Fig.~\ref{fig:mouselab}). As each click has a cost, finding the best possible path is a non-trivial problem. According to the theory, by collecting a sequence of clicks people made to decide which path is best, we gain an insight into people's planning process and the strategies they used.

\subsection{Approaches to improving human decision-making}\label{sec:boosting}
Interventions designed to tackle decision-making biases include educating people about rational decision-making \citep{larrick2004debiasing} and boosting people's decision-making by conveying simple heuristics they can use to arrive at good decisions \citep{hertwig2017nudging, hafenbradl2016applied}. In the first approach, harmful biases are tackled through motivating or incentivizing people to make better decisions by showing them what they could gain by adopting the principles of logic, probability theory, and expected utility theory. The key limitation of this approach is that those principles place unrealistically high cognitive demands on people when they are applied to nontrivial real-world problems \citep{larrick2004debiasing}. In response to the shortcomings of the first approach, the second approach directs people to use adaptive simple heuristics  \citep{hertwig2017nudging, hafenbradl2016applied,Gigerenzer1999}. These heuristics take advantage of common properties of decision problems, and allow the decision-maker to quickly make advantageous choices while being rather straightforward to understand. Two ways of conveying adaptive heuristics have been explored so far: teaching decision strategies and designing decision aids \cite{hafenbradl2016applied,hertwig2017nudging}.  

\subsection{AI-powered Boosting}\label{sec:aiboosting}
The general approach we followed in this study, called AI-powered boosting, is to leverage artificial intelligence (AI) to improve on previous efforts to boost human decision-making. The essence of AI-powered boosting is to employ machine learning to discover decision-making heuristics automatically and then convey them to people \citep{CognitiveTutorsPNAS,skirzynski2020automatic}. This approach rests on defining efficient decision strategies and computing them by solving appropriate optimization problems. According to this approach, optimal strategies for human decision-making do not attempt to maximize the expected outcomes (cf. Definition~\ref{deff:reward}) because that would be intractable for people. Computing the optimal decisions for every possible situation is simply too computationally demanding for both people and computers \citep{Simon1997,vanRoij2008}. Optimal decision strategies are instead defined as decision procedures that achieve the highest possible level of \emph{resource-rationality} \citep{lieder_griffiths}, which is defined as the expected utility of the choices that a given heuristic will make when a person uses it in a given environment  minus the opportunity cost of the time and mental resources it expends to reach those decisions.

When faced with a new decision, people generally cannot compute the resource-rational decision strategy for that scenario themselves, and doing so would be more demanding than computing the optimal decision \citep{lieder_griffiths,lieder2020advancing,rich2020intractability}. However, many resource-rational strategies can be executed with minimal effort once they have been learned \citep{CognitiveTutorsPNAS,He2021measuring,He2022Where,lieder_griffiths,lieder2020advancing}. Moreover, people can learn to efficiently detect which of their strategies is best suited for the decision they are facing \citep{lieder2017strategy}. This makes deriving resource-rational strategies for specific decisions people frequently face and conveying them to people a promising approach to improving human decision-making \citep{CognitiveTutorsPNAS}. 
Researchers interested in improving human decision-making can derive resource-rational strategies by leveraging artificial intelligence to compute the optimal policies of metalevel Markov Decision Processes \citep{Lieder2017,callaway2018learning,Griffiths2019}. Having solved such a problem, the result is a function for choosing the next planning operation according to information revealed by previous planning operations, called a \textit{metalevel policy}. 

Formally, AI-powered boosting \citep{CognitiveTutorsPNAS} proceeds by modeling the problem of selecting an optimal sequence of cognitive operations as a Markov Decision Process \citep{Sutton2018}, leverages reinforcement learning to derive the optimal policy for selecting planning operations from that model, and then conveys this policy to people. There are two different approaches for doing so: via intelligent cognitive tutors \citep{CognitiveTutorsPNAS,consul2021improving} and via AI-generated decision aids \citep{skirzynski2020automatic} (see Fig.~\ref{fig:SD}). Intelligent cognitive tutors train people to internalize the metalevel policy so that they can subsequently apply the automatically discovered decision strategy independently without further assistance. By contrast, AI-generated decision aids display a description of the automatically discovered decision strategy while people make their decisions.

AI-powered boosting has been shown to be a promising approach for improving human planning skills \citep{CognitiveTutorsPNAS}. The intelligent tutors presented in \citeauthor{CognitiveTutorsPNAS} (\citeyear{CognitiveTutorsPNAS}) taught people how to efficiently plan in the environment shown in Fig.~\ref{fig:mouselab}. After each planning operation performed by participants (i.e., a click uncovering a number), the tutor either praised them for following the optimal planning strategy, or penalized them by forcing to wait an amount of time proportional to the suboptimality of their planning operation. This procedure significantly improved people's planning skills when they interacted with a different, transfer environment. 
Subsequent work employed intelligent cognitive tutors that teach people by showing video demonstrations of the automatically discovered strategy \citep{consul2021improving,RSDArticle}. \citeauthor{consul2021improving} (\citeyear{consul2021improving}) developed intelligent cognitive tutors that improved people's planning in very large versions of the environment illustrated in Fig.~\ref{fig:mouselab}. 

Here, we extend and evaluate a recent AI-powered boosting method that conveys automatically discovered strategies through automatically designed decision aids \citep{skirzynski2020automatic}. The original version of this method comprises four steps. 

The first two steps are the same as in \cite{CognitiveTutorsPNAS}: the problem of selecting an optimal sequence of planning operations is modelled as a Markov Decision Process (MDP), and then reinforcement learning algorithms are utilized to find the optimal strategy from this model. This is where the first input to AI-powered boosting mentioned in the Introduction is required: the model of the decision problem. Here, we model decision problems in the Mouselab-MDP framework \citep{CallawayLiederKrueger2017,Callaway2022Rational}. This is a very general framework for modeling problems which involve sequentially processing, integrating, and selecting multiple pieces of information. This makes our approach applicable to a wide range of decision problems.

\begin{figure}[h!]
    \centering
    \includegraphics[width=0.7\linewidth]{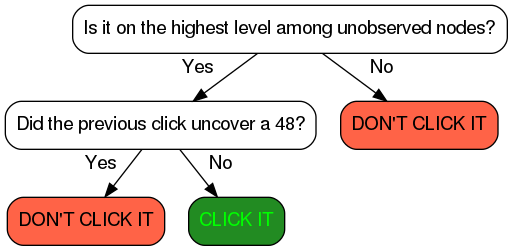}
    \caption{Sample flowchart obtained via the AI-powered boosting intervention from \citep{skirzynski2020automatic}. The flowchart expresses a far-sighted planning strategy in the Mouselab-MDP task with increasing variance}
    \label{fig:flowchart}
\end{figure}

In the third step, a set of logical primitives is created, and an imitation-learning algorithm called AI-Interpret uses these primitives to generate an interpretable description of the optimal strategy as a logical formula. Importantly, the formula generated by AI-Interpret is expressed in conjunctive normal form, that is a disjunction of conjunctions, which can be naturally expressed as a decision tree. Due to that, after the outputted description is automatically translated to natural language via a predefined primitives' dictionary in the fourth step, it is naturally formed into a static flowchart (see Fig.~\ref{fig:flowchart}). This flowchart evaluates to 1 (perform) or 0 (do not perform) for each action and each possible state people may encounter, and guides people through decision-making. The fourth step also introduces the second input to AI-powered boosting mentioned in the introduction: a dictionary of the relevant problem components. Our previous work \citep{skirzynski2020automatic} introduced a set of logical predicates capable of representing multiple planning strategies \citep{skirzynski2021automatic} used in the Mouselab-MDP, and hence in multiple planning problems that can be represented in this formalism. By translating these predicates into the elements of the problem (e.g. graph-theoretic \texttt{is\_leaf} predicate could mean long-term consequences of actions), our approach generates problem-specific natural language instructions instead of logical formulas. Like previously, successfully fulfilling this step relies on the researcher's knowledge of the problem and understanding of the Mouselab-MDP paradigm. For more complex problems, such as chess playing, it is also possible that new predicates need to be created and then translated.

Still, despite the success of AI-powered boosting in improving human decision-making through static descriptions \citep{skirzynski2020automatic}, we hypothesized that more naturalistic tasks may require a more easily interpretable representation of the decision strategy. In the next section, we present an experiment suggesting that procedural instructions are more effective than the decision aids constructed by the original version of boosting presented by \citet{skirzynski2020automatic}.

\section{Experiment 1: Procedural descriptions of planning strategies are more interpretable than static descriptions}
A key challenge of AI-powered boosting is to convey an abstract and potentially complex planning strategy to people. To support people effectively, the description of the strategy should be easy to understand. The description presented by the decision aid can either be \textit{static} or \textit{procedural}.
We call a family of descriptions that express a (planning) strategy by listing conditions under which certain actions can be taken \emph{static} descriptions. A decision tree is one example of a static description, and a decision set (a set of rules) is another example. In both cases, a sequence of conditions that describe the environment tell the decision maker when it is allowed to take which actions. A decision tree is the most extreme instance, since the conditions may be manifold, but specific actions have to be always coded inside the tree, whereas the actual actions either perform (1) or do not perform (0) the planning operation under consideration (see Section~\ref{sec:aiboosting}). In contrast to static descriptions, one can define \emph{procedural} descriptions that express (planning) strategies in terms of an ordered list of actions that need to be taken sequentially according to the order. A program is a primary example of a procedural description where the subroutines specify the actions, and their positioning in the program (top to bottom) defines the ordering. 

The intervention introduced in \citeauthor{skirzynski2020automatic} (\citeyear{skirzynski2020automatic}) made an important hidden assumption about the preferred way of conveying planning strategies to people, choosing static descriptions as the output. However, a more intuitive assumption is that procedural descriptions facilitate teaching planning strategies to a greater extent. If this were the case, then it might be possible to make AI-powered boosting substantially more effective by generating procedural descriptions rather than static descriptions. To find out if it would be worthwhile to develop algorithms for generating procedural instructions, we first tested whether procedural descriptions of planning strategies are more interpretable than static descriptions.

To do so, Experiment~1 was designed to determine if participants are more successful in applying a planning strategy when this strategy is expressed in terms of procedural instructions or the static descriptions used by \citeauthor{skirzynski2020automatic} (\citeyear{skirzynski2020automatic}). To answer this question, participants in Experiment~1 received decision aid on how to solve a sequential decision problem. One half of the participants received the static descriptions generated by the original version of the method \cite{skirzynski2020automatic} and the other half received hand-crafted procedural instructions. The main goal of Experiment 1 was to compare people’s ability to follow procedural versus static instructions. We therefore asked participants to execute the described strategy as accurately as possible. As a result, group differences in the task performance should result primarily from differences in the comprehensibility of static versus procedural descriptions.

\subsection{Methods}

\subsubsection{Task}
The template for the experimental task was the Mouselab-MDP paradigm \citep{CallawayLiederKrueger2017,Callaway2022Rational}. In this task, participants have to choose between six possible paths, each of which involves a series of 3 steps. People can gather information about how much reward they would receive for visiting a location by clicking on it for a small fee. The objective is to balance expenses while finding a rewarding path. \citeauthor{skirzynski2020automatic} (\citeyear{skirzynski2020automatic}) demonstrated that automatically generated static descriptions of planning strategies can boost participants' performance across multiple versions of that task. We adopted the experiment with a static description for the most demanding strategy for our purposes.

\begin{figure}[h!]
\begin{center}
\includegraphics[width=0.99\linewidth]{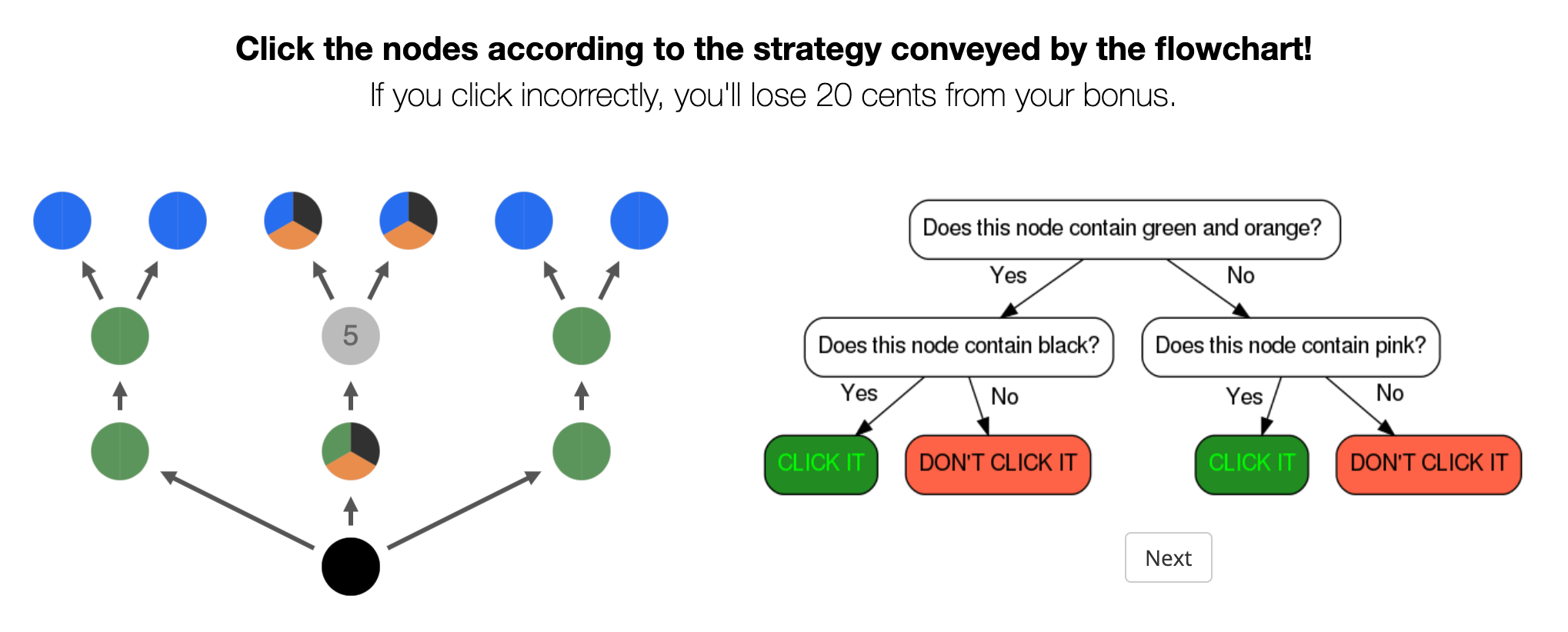}
\end{center}
\caption{Experiment 1: The experimental screen as shown to participants in the flowchart condition. On the right, the Mouselab-MDP task with color coded nodes. Each color represents a node property, which can change depending on the uncovered values. On the right, the color coded flowchart depicting a specific clicking procedure. In the alternative condition the flowchart is replaced with procedural instructions}
\label{fig:exp1_task}
\end{figure}

To create an unbiased comparison of static descriptions and procedural instructions, we created an experiment where the participants' only task is to execute the described strategy. Participants were instructed to select planning operations (i.e., clicks) according to the described strategy. To minimize the risk that participants misunderstood the task as collecting rewards, participants were not informed that the numbers correspond to rewards and could neither pick a path nor collect any rewards. This resulted in a sequential clicking task in which participants' only objective was to click nodes according to presented instructions. The original static descriptions described the locations to click based on attributes derived from graph theory, for example whether a location is on the current most rewarding path and whether a location can be found at a certain depth of the graph. These kinds of descriptions can be difficult to understand for laymen and usually require training. To simplify the procedure, each attribute was visually presented as a color. Based on the uncovered rewards, we computed if an attribute applied to a location and, if so, the location was marked with the corresponding color (see Fig.~\ref{fig:exp1_task}). One location could be tagged with multiple colors. The attributes used in the static descriptions and in the procedural instructions were replaced by the corresponding color names. 
The manipulations described had two main advantages. First, participants were not tempted to neglect the clicking instructions and apply their own clicking strategy, as we concealed the original planning task. Second, the instructions required only knowing the colors, giving very little room for misinterpretations compared to more complex attributes in the original task. In sum, this manipulation eliminated possible limitations of the original task and let us compare the degree of interpretability of both static descriptions and procedural instructions more directly. 
The environment participants interacted with was the color-coded Mouselab-MDP and was displayed on the left of the experiment's screen, whereas the instructions appeared on the right of the screen (see Fig.~\ref{fig:exp1_task}). The static descriptions were the same as the ones used in \cite{skirzynski2020automatic} except that the attributes were now color-coded. The procedural instructions were hand-crafted and described the same clicking strategy as the static descriptions. The instructions were as follows: ``Click the nodes that contain green and orange until you find a +10. Then click the nodes that contain blue and pink.''.

\subsubsection{Dependent variables}\label{sec:exptwodv}
To measure whether a participant understood the instructions, we counted the number of clicks that were consistent versus inconsistent with the strategy they were instructed to follow. Both measures were combined into one metric called click agreement, which we defined as the proportion of consistent clicks out of all performed clicks, that is \begin{equation} 
\text{agreement} = \frac{n_{\text{consistent}}}{n_{\text{consistent}} + n_{\text{inconsistent}}}.
\end{equation}
In the event that a participant finished the trial before having clicked all the instructed locations, the expected number of missed clicks counted towards the inconsistent clicks. We calculated the number of missed click as the average number that the instructed strategy performed in 1000 simulations minus the number of clicks performed by the participant. Click agreement is reported in percent. In addition, we measured the expected value of the score participants would have received if they had performed their clicks in the Mouselab-MDP task. By definition, the expected score is the sum of rewards along the best path identified by the participant's clicks minus the cost of those clicks. Because the expected value of unrevealed rewards was zero, the expected reward of a path connecting the start node was equal to the sum of the rewards revealed on that path. The cost of the participant's clicks was $1$ point per click.

\subsubsection{Procedure}
Participants were randomly assigned to a static descriptions or a procedural instructions condition. Both conditions started with an instructions block, in which the task was introduced and motivated as a pass-code test required to enter an extraterrestrial planet that can only be accessed by following specific instructions. This was followed by an attention check consisting of 3 multiple-choice questions, a block explaining the procedure and the instructions (see Section~\ref{appendix:exponeTaskInstructions}) and a final attention check consisting of 3 multiple-choice questions. After this, participants engaged in a single practice trial in which they were instructed to click locations marked with orange. They received feedback on the correctness of their clicks and when to end the trial. Lastly, there was a block of 10 test trials and a small demographic survey. The minimum time required to spend on a trial was ten seconds. Participants were informed that they would start with a bonus of \$2 and lose 20 cents for each trial in which they made an inconsistent click or quit early. Neither their current amount of bonus nor the score was displayed in the task. In addition, all participants received a base payment of \$1.

\subsubsection{Participants}
We recruited 21 people for the static descriptions condition and 21 people for the procedural instructions condition on Amazon Mechanical Turk (average age: 35.9 years, range: 18-65 years; 21 female). The experiment lasted 11.9 minutes on average. Our predefined exclusion criterion, which excludes individuals who do not perform any click in half of the test trials, did not apply to any participant.

\begin{figure}[h!]
\begin{center}
\includegraphics[width=1\linewidth]{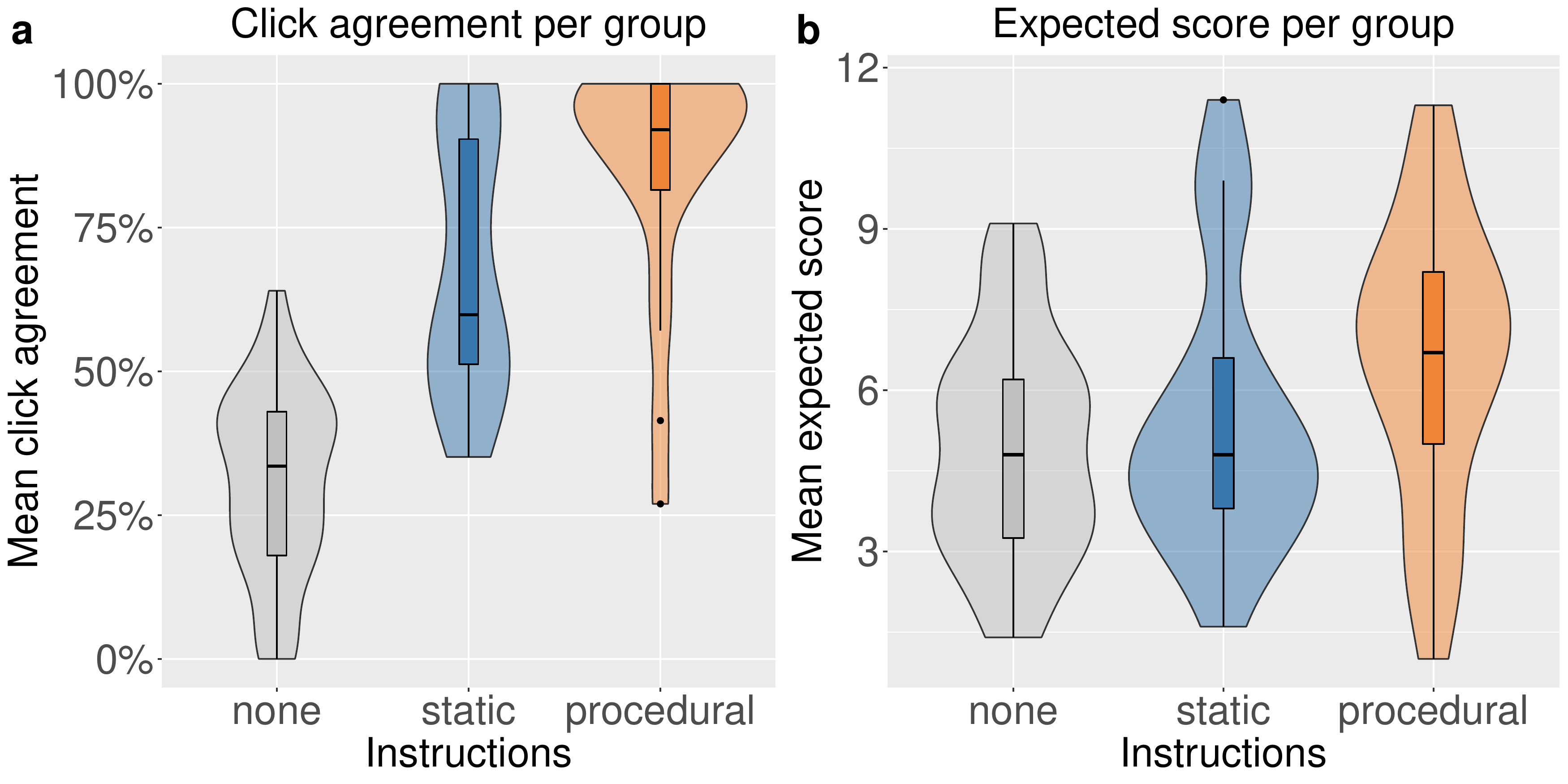}
\end{center}
\caption{Experiment 1: a) Procedural instructions are more interpretable than static descriptions (\textit{p}=.018). The plot shows the mean click agreement of participants of the experimental conditions (blue and red) and a control condition (grey).  b) Participants assisted by procedural instructions had higher expected scores than participants assisted by static descriptions (\textit{p}=.09). The expected score that participants would have received if they had performed their clicks in the Mouselab-MDP task
}
\label{fig:exp1_results}
\end{figure}

\subsection{Results}
The mean click agreement was 68.8\% (Median: 59.8\%, \textit{SD}=22.9\%) in the static descriptions condition and 85.2\% (Median: 92.0\%, \textit{SD}=20.5\%) in the procedural instructions condition, as illustrated in Fig.~\ref{fig:exp1_results}a. The variable was not normally distributed; we thus employed a one-sided Mann-Whitney-U test. We found that click agreement was significantly higher in the procedural instructions condition ($A=.69$\footnote{For non-parametric tests, we report the common language effect size \textit{A}, which describes the probability that a randomly chosen member of group 1 scores higher on the dependent variable than a randomly chosen member of group 2 \citep{ruscio2008probability}}; \textit{U}=137, \textit{p}=.018). Click agreement and expected score were positively correlated (\textit{r}(40)=.45 \textit{p}=.003). Accordingly, the mean expected score was higher in the procedural instructions condition (\textit{M}= 6.6, \textit{SD}=2.7; see Fig.~\ref{fig:exp1_results}b) than in the static descriptions condition (\textit{M}= 5.5, \textit{SD}=2.6), although this difference was not statistically significant (\textit{d}=0.41; \textit{t}(40)=-1.3, \textit{p}=.09). The learning curves for click agreement and expected score over the ten trials can be found in the Appendix (see Section~\ref{fig:exp1_learningcurves}). Furthermore, participants assisted by procedural instructions were able to reach their decisions significantly faster than participants assisted by static descriptions (21.0~sec vs. 31.6~sec; \textit{d}=0.95, \textit{t}(40)=3.1, \textit{p}=.002). Moreover, we compared the performance of the two experimental groups to a control group ($N=60$) that was instructed to maximize their score in the Mouselab-MDP task without being assisted by a decision aid\footnote{The control condition was part of Experiment 3 of our previous work (Skirzyński et al., 2021)}. The click agreement in both, the static descriptions condition ($A=.93$; \textit{U}=93, \textit{p}=<.001) and the procedural instructions condition ($A=.95$; \textit{U}=59, \textit{p}=<.001) was significantly larger than in the control condition (\textit{M}= 31\%, \textit{SD}=16\%). This also applied to the expected score which was significantly larger in the static descriptions condition (\textit{d}=0.49, \textit{t}(79)=1.9, \textit{p}=.027) and the procedural instructions condition (\textit{d}=0.9, \textit{t}(79)=3.6, \textit{p}=.<001) than in the control condition (\textit{M}= 4.2, \textit{SD}=2.6).

\subsection{Discussion}
The experiment was designed to compare people’s ability to follow procedural versus static instructions on how to make a plan. We deliberately reduced the potentially confounding effect of participants’ general decision-making competency, domain knowledge, insights into the specific decision problem, and opinions about which strategy might be best. To achieve this, we chose a highly abstract task and instructed all participants to follow the instructions as accurately as possible. We can therefore interpret the observed differences in task-performance primarily in terms of the comprehensibility of the instructions.
In accordance with our hypothesis, we found that procedural instructions were more helpful for people than static descriptions. Even though the final expected score obtained in the experiment did not differ significantly between the conditions, the group which utilized the procedural instructions was both faster and numerically more accurate in applying the described strategy. The main reasons why the noticeable numerical difference in accuracy was not statistically significant might be that our sample size was rather small relative to the high variance of the rewards in this particular task. The comparison to a control group showed that both decision aids significantly improved participants' adherence to the instructed strategy and the corresponding gain in expected score. 
One possible explanation for the results we observed is that static descriptions underutilize people’s capacity to comprehend and execute structured, abstract procedures \citep{miller1960plans}. This interpretation suggests that the static descriptions were unnecessarily detailed, long, and repetitive.

We acknowledge that the optimal strategies for some difficult problems could be so complex that they cannot be approximated well by any verbal description. Apart from those cases, the results of Experiment 1 should also hold for more complex tasks because the difference in the amount of effort required to follow static versus procedural instructions would be even larger for larger tasks. People's increased compliance with the near-optimal strategy and their faster decisions found in Experiment~1 thus suggest that it is worthwhile to extend our AI-based boosting method to procedural instructions because people appear to be much more willing and able to follow them.

\section{AI-powered boosting with procedural descriptions of optimal decision strategies}
As mentioned in Section~\ref{sec:aiboosting}, the original version of AI-powered boosting \citep{skirzynski2020automatic} generates static descriptions in form of a flowchart (decision tree) for verifying whether a candidate planning operation is consistent with an optimal planning strategy. A key limitation of this approach is that it does not explicitly tell people which planning operations to perform but requires them to come up with a good planning operation themselves and then verify their hypothesis. Motivated by the results of Experiment~1, we now present an algorithm for transforming the output of AI-Interpret into a procedural description of how to plan that explicitly states which planning operation should be performed first, second, third, and so on. 
By coupling AI-Interpret with this new algorithm, we obtain a general method for describing any RL policy through procedural instructions. Our method is very general because AI-Interpret is a policy-agnostic method that only utilizes demonstrations of the policy to describe it. It therefore contributes not only to the field of boosting human decision making but also to the field of explainable reinforcement learning \citep{puiutta2020explainable, dazeley2021explainable}. Notably, both AI-Interpret \citep{skirzynski2020automatic} and our new algorithm compute approximations to the optimal policy and the flowchart, respectively. AI-powered boosting hence generates lossy simplifications of the policies computed by dynamic programming or reinforcement learning. However, those simplifications are not only easier to grasp than static descriptions (see Experiment~1), but they also achieve their objective to improve human decision-making, as will we show in the following sections.

Later in this article, we will focus on the application of our new method to AI-powered boosting rather than standard RL tasks. This section hence presents our two major technical contributions. In the first subsection, we detail DNF2LTL -- our algorithm for transforming disjunctive normal form logical formulas into procedural descriptions. In the second subsection, we present AI-powered boosting extended with this algorithm.

\subsection{Generating procedural descriptions of planning strategies}
The original output of AI-Interpret that is utilized by AI-powered boosting is a Disjunctive Normal Form formula (DNF; see Definition~\ref{def:dnf}). Our algorithm, which we call DNF2LTL, transforms such formulas into the procedural format of Linear Temporal Logic (LTL; see Definition~\ref{def:ltl}). DNF2LTL operates in two phases. In the first phase, it modifies the input disjunctive normal form formula into an entity called a \emph{procedural formula} (see Definition~\ref{def:procf}), that is an expression in a specific from of linear temporal logic. In the second phase, the procedural formula is pruned to remove as many unnecessary predicates as possible to obtain the possibly simplest procedural description of the automatically discovered strategy. We formally define the output of our algorithm in the next section.

\subsubsection{Procedural formulas}
Procedural formulas generated by our algorithm are special cases of Linear Temporal Logic (LTL) formulas extended with two additional operators. 

First, LTL itself is a type of propositional logic that allows expressing processes that change in time (see Definition~\ref{def:ltl}).
\begin{deff}[Linear Temporal Logic]\label{def:ltl}
Let $\mathcal{P}$ be the set of propositional variables $p$ (variables that can be either true or false), let $\lnot, \land, \lor$ be standard logical operators for negation, AND, and OR, respectively, and let $\mathbf{X}, \mathbf{U}, \mathbf{W}$ be modal operators for NEXT, UNTIL, and UNLESS, respectively. Linear temporal logic (LTL) is a logic defined on (potentially infinite) sequences of truth-assignments of propositional variables. LTL formulas are expressions that state which of the variables are true, and when they are true in the sequence. Whenever this agrees with the actual truth-assignment in an input sequence, then we say that a formula is true. 

Formally, for $\alpha$ and $\beta$ being LTL formulas, we define a formula to be expressed in LTL inductively: $\psi$ is an LTL formula if $\psi\in\mathcal{P}$ ($\psi$ states that one of the variables is true in the first truth-assignment in the sequence), $\psi= \lnot\alpha$ ($\psi$ is a negation of an LTL formula), $\psi=\alpha\lor\beta$ ($\psi$ is a disjunction of two LTL formulas), $\psi=\alpha\land\beta$ ($\psi$ is a conjunction of two LTL formulas), $\psi = \mathbf{X} \alpha$ ($\psi$ states that LTL formula $\alpha$ is true starting from the next truth-assignment in the sequence) or $\psi = \alpha \mathbf{U} \beta$ ($\psi$ states that LTL formula $\alpha$ is true until some truth-assignment in the sequence where LTL formula $\beta$ becomes true). 
\end{deff}
We extend the standard definition of LTL to allow more natural transition from decision trees to procedural instructions that, as we found in Experiment~1, are easier for people to follow. To do so, we add a Hold operator that allows introducing a default stopping condition (see Definition~\ref{deff:holding}), and a Loop operator that defines which part of the procedure to repeat (see Definition~\ref{deff:loop}).
\begin{deff}[Hold modal operator]\label{deff:holding}
The HOLD operator $\mathbf{H}$ is a unary operator in the linear temporal logic. LTL formula $\mathbf{H}\psi$ states that $\psi$ is true at least for the first truth-assignment in the sequence of truth-assignments, and then eventually becomes false. HOLD operator is the UNTIL operator with a default until condition ("until it is no longer satisfied").
\end{deff}

\begin{deff}[Loop modal operator]\label{deff:loop}
The LOOP operator $\mathbf{L}$ is a binary operator in the linear temporal logic. LTL formula $\phi \mathbf{L} \psi$ states that i) part of $\phi$ is an LTL formula $\psi$, ii) if $\psi$ is replaced with some number of NEXT $\psi$ operators $\mathbf{X} \psi \land\dots\land\ \mathbf{X} \psi$, then the new $\phi$ is true across the whole sequence of truth-assignments. In other words, the LOOP operator states that in order for LTL formula $\phi$ to be true and satisfy all truth-assignments in the sequence, LTL formula $\psi$ inside of $\phi$ needs to be repeated the appropriate number of times (form a ``loop''). This formula corresponds to truth-assignments in the sequence that disagree with $\phi$.
\end{deff}
Finally, by adding the introduced operators to the LTL formalism, we obtain procedural formulas (see Definition~\ref{def:procf}).
\begin{deff}[Procedural formula]\label{def:procf}
We say that $f$ is a procedural formula if and only if $f$ is an expression written in linear temporal logic where the propositional variables are predicates $h: \mathcal{S}\times\mathcal{A}\rightarrow\{0,1\}$ for some set of states $\mathcal{S}$ and some set of actions $\mathcal{A}$, and where the modal operators are $\mathbf{X}$ (NEXT), $\mathbf{U}$ (UNTIL), $\mathbf{W}$ (UNLESS), $\mathbf{H}$ (HOLD), and $\mathbf{L}$ (LOOP).
\end{deff}

\subsubsection{Transforming disjunctive normal form formulas into procedural formulas}
In the first phase, DNF2LTL generates a procedural formula (see the previous section) out of a DNF formula (see Definition~\ref{def:dnf}). 
\begin{deff}[Disjunctive Normal Form]\label{def:dnf}
Let $f_{i}, h: \mathcal{X}\rightarrow\{0,1\}$ for $i\in\mathbb{N}$ be binary-valued functions (predicates) on domain $\mathcal{X}$. We call $f_{1}(\boldsymbol{x}) \lor f_{2}(\boldsymbol{x}) \lor \dots \lor f_{n}(\boldsymbol{x})$ a disjunction of predicates and $f_{1}(\boldsymbol{x}) \land f_{2}(\boldsymbol{x}) \land \dots \land f_{n}(\boldsymbol{x})$ a conjunction of predicates. We say that $h$ is in disjunctive normal form (DNF) if $h$ is a conjunction of disjunctions of predicates $f_{i}$.
\end{deff}
To do so, our algorithm accepts four main inputs: the set of trajectories that led to the creation of the DNF formula, a set of predicates that could serve as the until or unless conditions, a set of predicates which are unwanted in the procedural formula, and, naturally, the DNF formula itself. 
\begin{deff}[Trajectory]\label{def:traj}
A trajectory $\tau = [\phi_0,...,\phi_{N-1}]$ is a sequence of $N$ state-action pairs $\phi_i = (s_{i},a_i)$ with $s_{i}\in\mathcal{S}, a_i\in\mathcal{A}, i=0,...,N-1$. 
\end{deff} 
The trajectories (see Definition~\ref{def:traj}) play the role of the sequences of truth-assignments from Definition~\ref{def:ltl}, whereas the set of predicates for until/unless conditions and the DNF formula define the building blocks out of which the procedural formula would be constructed. The other remaining parameter is optional, and in case of a failure in producing an output, the algorithm is ran again without removing the redundant predicates. On a high level, our algorithm exploits the idea that a DNF formula is satisfied when at least one of its conjunctions is satisfied. It iterates over the trajectories to discover the dynamics of changes in truth values of the conjunctions, and uses the conjunctions, the found dynamics, and the candidate until/unless conditions to generate procedural formulas. 

During the first phase, DNF2LTL generates an initial procedural description in four steps. In the first step, the algorithm extracts potential subroutines from the inputted DNF formula. In the second step, the algorithm determines the order in which those subroutines should be performed. In the third step, the algorithm computes the logical conditions for transitioning from each step to the next. Finally, in the fourth step, our method connects the subroutines with the appropriate conditions into a complete procedural description and outputs the result. Algorithm~\ref{dnf2ltl} presents a pseudocode that implements the first phase of DNF2LTL and the following paragraph provides a technical description of each of these four steps in greater detail. We relate this description to the pseudocode by listing its relevant line numbers in brackets. Readers who are primarily interested in the big picture and the application to boosting human planning can skip these technical details.
\begin{enumerate}[label=Step \arabic*: ]
    \item DNF2LTL starts by dividing the DNF formula into a set of conjunctions and removing all the unwanted predicates [Lines 3-4]. 
    \item Then, it iterates over the trajectories and for each trajectory records the sequence of conjunctions that were true for that trajectory so that the whole DNF formula could be true across all the state-action pairs within it. Our algorithm then creates a transition graph where conjunction $c_i$ is connected with conjunction $c_j$ if there is at least one trajectory$\tau$ where the value of $c_i$ changed from true to false at the same moment when the value of $c_j$ changed from false to true [Line 9]. The transition graph is used to generate maximum length sequences of conjunctions $c_{i_1}c_{i_2}\dots c_{i_n}$ to capture the possibly fullest transition evidenced in the data [Line 10]. The last predicate in this sequence (i.e., $c_{i_n}$) either has no outgoing connections in the transition graph or connects to one of the $c_{i_j}$s in which case the sequence ends with a special loop symbol that indicates which $i_j$ that is. The resulting maximum length sequences are used to define equivalence classes for the trajectories. These equivalence classes represent potential dynamics of how the conjunctions change their truth values so that the full DNF formula was satisfied. Each trajectory, treated as a sequence of conjunctions of the DNF formula, is then assigned to a number of equivalence classes. Namely, trajectory $\tau$ represented by sequence $s$ is assigned to all equivalence classes $e$ for which $s$ is a subsequence of $e$. For instance if $\tau$ is represented by sequence $c_1c_3$, it could be assigned to equivalence class $c_1c_2c_3c_4\ \text{LOOP}\ c_2$ [Line 11]. This whole process in Step 2 is performed to generate candidates for procedural descriptions.
    \item Then, the algorithm transforms unempty equivalence classes into procedural formulas. It does so by using the trajectories in the class [Line 13] to iteratively find UNTIL operators (UNTIL conditions) that could separate each of the elements in the sequence representing the class. During one iteration, DNF2LTL searches for the UNTIL condition separating a pair of subsequent conjunctions. Possible candidates for UNTIL conditions are the allowed predicates provided as an input to the method and 2-element disjunctions of those predicates [Input P], i.e. we hand-engineer possible operators a priori. For a pair of subsequent conjunctions $c_ic_{i+1}$, the matching conditions are such whose truth value changes from constantly false, while $c_i$ is true, to true, when $c_{i+1}$ true. We select the UNTIL condition among matching conditions as the $u_i$ that maximizes the likelihood of the trajectories in the equivalence class under $c_1\ \text{UNTIL}\ u_1\ \text{AND NEXT}\ \dots c_i\ \text{UNTIL}\ u_i\ \text{AND NEXT TRUE}$, i.e. the formula generated so far [Lines 21-24]. This process allows us to select a condition that we know is appropriate (belongs to input P), and that is the most likely under the data. If there is no matching condition, the algorithm adds the default UNTIL condition -- the HOLD operator or, if some predicates were removed from the formula, tries again with the original formula [Lines 25-29]. If some trajectories have a conjunction representation shorter than the representation of the class, the algorithm also adds an UNLESS operator after the UNTIL operator, and searches for the UNLESS condition in a similar way. If there are no matching conditions in input P, $FALSE$ is selected as the condition so that to allow excessive planning and prevent UNLESS to be met. This process models a situation when the formula allows early stopping [Lines 29-40]. Having set the condition(s), the generated LTL subformula is attached to the formula built so far via the NEXT operator [Lines 42-44]. The algorithm then iterates [Line 15]. If the last pair of conjunctions from the sequence representing the equivalence class contains a conjunction and a loop symbol, this symbol is transformed into the LOOP operator (see Definition~\ref{deff:loop}) and the conjunction and the loop operator are joined through the NEXT operator [Lines 15-20]. If there are demonstrations that end before the loop, the UNLESS operator is added in the same way as before [Line 19].
    \item After generating the procedural formulas for each of the equivalence classes, the final procedural formula is returned as a disjunction of these formulas [Line 49]. Note, however, that only one of the elements in the disjunction is returned by DNF2LTL after it performs pruning (see below).
\end{enumerate}

Our algorithm captures a special type of procedural formulas. For a DNF formula with only one conjunction, the structure of the output can be described by the regular expression
\begin{equation}\label{eq:proc}
    \left[\mathbf{H}\ \Phi \land \mathbf{N}\ |\ \Phi\ \mathbf{U}\ P\ (\mathbf{W}\ P) \land \mathbf{N}\right]^+\left[\mathbf{H}\ \Phi \ |\ \Phi\ \mathbf{U}\ P\ |\ \mathbf{L}\ P\right](\mathbf{W}\ P)
\end{equation}
where $P$ may be substituted with either of the input allowed predicates or their 2-element disjunction, and $\Phi$ may be substituted with an arbitrary conjunction of those predicates. The expression given in Equation~\ref{eq:proc} thus generates procedural formulas in the form of a sequence of NEXT operators, where subsequent conjunctions are separated with UNTIL conditions (and/or UNLESS conditions) or accompanied by the HOLD operator. The formula ends with the last NEXT operator or with a LOOP operator.

\begin{algorithm}
\SetAlgoLined
\SetKwInput{KwInput}{Input}
\SetKwInput{KwOutput}{Output}
\KwInput{DNF formula $f$;}
\myinput{Set of $N$ trajectories $D = \{((s_{1j},a_{1j}))_{j=1}^{l_1},\dots,((s_{Nj},a_{Nj}))_{j=1}^{l_N}]\}$;}
\myinput{Unless/until predicates $P$;}
\myinput{Redundant predicates $R$;}
\KwOutput{Procedural formula $\psi$;}
\begin{algorithmic}[1]
 \STATE Transform the trajectories in $D$ into a matrix of predicate truth-assignments $M_D$
 \STATE $L\leftarrow \text{lengths of trajectories in } D$
 \STATE Transform $f$ to a list of $m$ conjunctions $C=[p_{11} \land \dots \land p_{1n_1},\dots,p_{m1} \land \dots \land p_{mn_m}]$.
 \STATE Remove redundant predicates $R$ from $C$.
 \IF{Nothing was removed and $len(C) == 1$}
     \STATE $\xi = C[0]$
     \RETURN $\psi = \mathbf{H}\ \xi$
 \ENDIF
 \STATE $G\leftarrow \text{graph of connections between }c\in C \text{ based on } D$
 \STATE $S\leftarrow \text{max length sequences from }G$
 \STATE $E\leftarrow \text{equivalence classes from } S, D$
 \FOR{$e \in E$}{
   \STATE $e_{trajs}\leftarrow \text{Trajectories from } e$.
   \STATE $\beta_{prev}=\emptyset$
   \FOR{$\beta$ in $e$}{
     \IF{$loop$ in $\beta$}
         \STATE Use the special $loop$ symbol in $\beta$ to determine the conjunction to return to $\gamma$.
         \STATE $\phi = \phi\ \mathbf{L}\ \gamma$
         \STATE Go to step 29
     \ENDIF
         \STATE $\Theta\leftarrow \text{possible until conditions found with } e_{trajs}, M_D, \beta, \beta_{prev}$.
         \IF{success}
             \STATE $\theta^*\leftarrow \argmax_{\theta\in\Theta}\text{likelihood}(e_{trajs}, \phi \land \mathbf{N}\ \beta\ \mathbf{U}\ \theta)$
             \STATE $\phi = \beta\ \mathbf{U}\ \theta$
         \ELSIF{nothing was removed from $C$}
             \STATE $\phi = \mathbf{H}\ \beta$
         \ELSE
             \STATE Run the algorithm with $R=\emptyset$
         \ENDIF
         \STATE Determine if $\exists \tau \in e_{trajs}: len(\tau)<len(e)$ and the unless operator is necessary.
         \IF{necessary}
            \STATE $\Theta\leftarrow \text{possible unless conditions found with } e_{trajs}, M_D, \beta$.
             \IF{success}
                 \STATE $\theta^*\leftarrow \argmax_{\theta\in\Theta}\text{likelihood}(e_{trajs}, \phi\ \mathbf{W}\ \theta)$
                 \STATE $\phi = \phi\ \mathbf{W}\ \theta$
             \ELSIF{nothing was removed from $C$}
                 \STATE $\phi = \phi\ \mathbf{W}\ \mathbf{false}$
             \ELSE
                 \STATE Run the algorithm with $R=\emptyset$
             \ENDIF
         \ENDIF
         \IF{$\beta$ is not last in $e$}
             \STATE $\phi = \phi \land \mathbf{N}$
         \ENDIF
   }
   \ENDFOR
   \IF{$\beta$ is the first element in $e$}
       \STATE $\psi = \phi$
   \ELSE
       \STATE $\psi = \psi \lor \phi$
   \ENDIF
   \STATE $\beta_{prev}=\beta$;
 }
 \ENDFOR
 \RETURN $\psi$
 \end{algorithmic}
 \caption{First phase of DNF2LTL which generates procedural formulas out of DNF formulas.}
 \label{dnf2ltl}
 \end{algorithm}
 
\subsubsection{Pruning}
After our algorithm generates a procedural formula $\Psi$ in the first phase, it enters the second phase. During the second phase, DNF2LTL prunes the predicates appearing in the conjunctions of $\Psi$. Recall, however, $\Psi$ is a disjunction of procedural formulas. Because of this reason, pruning occurs for each element of that disjunction separately. To do so, DNF2LTL maps each procedural formula $\psi_i$ of the disjunction $\Psi$ onto a distinct binary vector $b_i$. Each element of $b_i$ is the truth value of one of the predicates appearing in the conjunctions making up $psi_i$. Our algorithm iterates over $b_i$s and in each step performs a greedy optimization. Concretely, for each consecutive predicate of $\psi_i$ the corresponding entry of  $b_i$ is set to zero if and only if removing that predicate increases the likelihood of the trajectories under the pruned description relative to the unpruned description. Some predicates increase the likelihood and are consequently pruned. After performing this optimization for each $b_i$ (and $\psi_i$), the algorithm outputs the pruned $\psi_i$ for which the likelihood was the highest as the final procedural description.

\subsection{Extending AI-powered boosting to procedural descriptions}
\begin{figure}
\begin{center}
\includegraphics[width=0.99\linewidth]{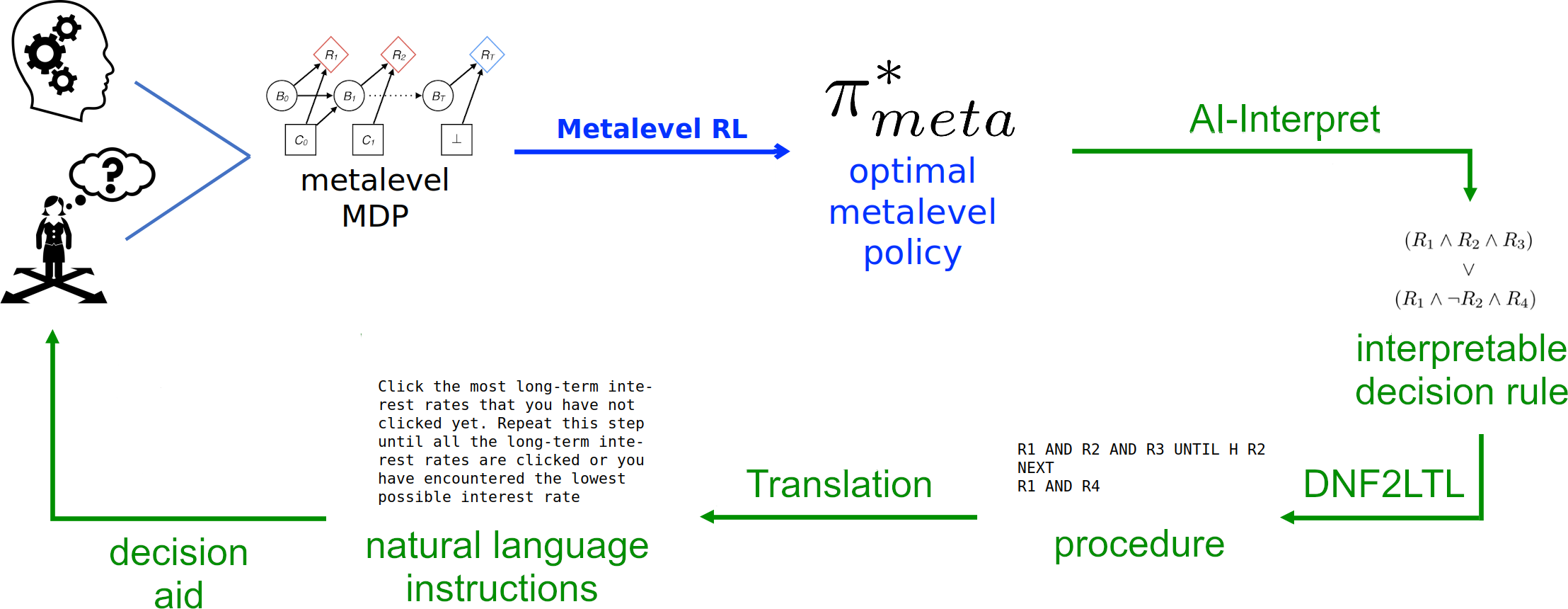}
\end{center}
\caption{AI-powered boosting with decision aids conveys automatically found planning strategies through automatically generated procedural instructions. Automation in the former area is achieved by modeling the problem as a metalevel MDP and solving it with metalevel reinforcement learning. Automation in the latter area rests on i) utilizing an imitation learning algorithm AI-Interpret that constructs a DNF formula of predefined predicates to describe the strategy, ii) applying the DNF2LTL method that transforms the DNF into a formula in LTL, iii) translating the output into natural langauge using a pre-defined predicate-to-expression dictionary}
\label{fig:isd}
\end{figure}

Following the result on human preference toward procedural descriptions of planning strategies, we extended the vanilla AI-powered boosting method by adding our DNF2LTL algorithm (see Fig.~\ref{fig:isd}). Instead of generating static, flowchart descriptions of optimal decision strategies, our extended version performs AI-powered boosting with procedural descriptions of optimal decision strategies. Originally, AI-powered boosting models the decision problem theoretically, finds the optimal policy for that model and relies on the AI-Interpret algorithm to find a static description of this policy. Lastly, it translates this description to natural language (see Section~\ref{sec:aiboosting}). In the new version of AI-powered boosting, we transform the description generated by AI-interpret into a program-like procedural description of logical primitives using DNF2LTL. We translate these descriptions into natural language instructions only after this transformation. In total, our new method comprises five steps: 1) modeling the planning problem, 2) finding the optimal strategy through that model, 3) creating a description of that strategy in form of a logical formula, 4) changing the formula to a procedural formula, 5) translating the procedural formula to natural language instructions.

To evaluate our extension of the general AI-powered boosting method, we applied it to discover and teach the optimal planning strategy for the three-step planning task illustrated in Fig.~\ref{fig:mouselab}. This planning task presents a version of the Mouselab-MDP paradigm (see Section~\ref{sec:mouselab_mdp}) created by \citeauthor{lieder2019cognitive} (\citeyear{lieder2019cognitive}) where the variance of the possible rewards is small in the first step (immediate consequences), becomes slightly larger in the second step (short-term consequences), and much larger in the third step (long-term consequences). In this scenario, people often neglect inspecting the long-term consequences of their actions \citep{jain2021computational, Callaway2022Rational, lieder2019cognitive}. By contrast, the optimal planning strategy for this environment takes a far-sighted approach to decision-making that would be beneficial in the real world. We hypothesized that our extended AI-powered boosting approach may represent this strategy in a number of more naturalistic decision problems, and help people improve their decision-making therein. 

In applying the extended AI-boosting method to new problems, we completed steps 1 to 4 using the same methodology and the same parameters as in \cite{skirzynski2020automatic}, but focused specifically on the Mouselab-MDP with increasing variance structure. In step 4 we selected candidate predicates for the until and unless conditions using the domain-specific language introduced in \cite{skirzynski2020automatic}. In step 5 we adapted the predicate-dictionary introduced in \cite{skirzynski2020automatic} to work on procedural descriptions and be domain-specific, depending on the problem that AI-powered boosting is to provide decision support on. Concretely, the procedural description is created by separating the formula into steps, where each step is a part of the formula between two NEXT operators (or the part between the beginning of the formula and the first NEXT operator or the part between the last NEXT operator and the end of the formula). The steps are then separately translated according to the alignment of LTL operators and predicates in the step, using the mentioned domain-specific dictionary. The whole description is returned as an enumeration of those translations.

Concretely, each step is translated according to the following logic: if there is at least one non-negated predicate in a step (other than $TRUE$), the translation always begins with the template ``\texttt{ACT }\emph{pred}\texttt{(OBJ, REW)}'', where \texttt{ACT} is substituted with a domain-specific action word, and \emph{pred}(\texttt{OBJ, REW}) is substituted with a domain-specific translation of predicate \emph{pred} that contains \texttt{OBJ} as the object word represented by nodes in the Mouselab-MDP, and \texttt{REW} as the reward word represented by numbers hidden underneath the nodes. For instance, the action word could be ``\emph{Look up}'', the object word could be ``\emph{hotels}'', the reward word could be ``\emph{the prices}'' and the predicate could capture the property of being positioned in the last level from the start node of the Mouselab-MDP. The translation would then read ``'\emph{look up the prices of the most distant hotels}''.

If there are any negated predicates in a step, they are translated afterwards. The translations of those predicates themselves are listed in bulletpoints following the template ``\emph{Do not }\texttt{ACT}\emph{:}''.
Special predicates include the always $TRUE$ predicate, which is translated to ``\emph{Stop planning right away or }\texttt{ACT}\emph{ some random }\texttt{OBJ} \emph{and then stop planning}'' (because the available planning operations always include both clicking the nodes and terminating), and always $FALSE$ predicate, which is translated using the template ``\emph{Do not }\texttt{ACT}\emph{ anything}'' (because the always $FALSE$ predicate appears in the description only on its own -- for the ``no-action'' strategy). 

For the translation of the temporal operators, when a step includes an \texttt{UNTIL} operator with condition \emph{cond(}\texttt{OBJ, REW}\emph{)}, the translation template becomes ``\emph{Repeat this step until cond(}\texttt{OBJ, REW}\emph{)}''. If there is an UNLESS operator with condition \emph{cond(}\texttt{OBJ, REW}\emph{)} in a step, the translation of the step starts with the text ``\emph{Unless cond(}\texttt{OBJ, REW}\emph{)}\emph{, in which case stop at the previous step}''. If the there is a HOLD operator in a step, the translation of the step is added the following text ``\emph{Repeat this step as long as possible}''. Finally, if there is a LOOP operator with expression EXPR in a step (which can only occur in the last step), the algorithm matches the number of the step in which expression EXPR appeared the latest, say \texttt{NUM}, and adds the template ``\emph{GOTO step} \texttt{NUM}'' to the translation. The exact translation for the predicates in our DSL and the rules governing how the steps are translated can be found in our project's repository in the \texttt{translation.py} file.

For our purposes, the final procedural formula we obtained was 
\begin{align}
\begin{split}
    & among(not(is\_observed, has\_largest\_depth)\ \mathbf{U}\ \\ & (are\_leaves\_observed\ \lor\ is\_previous\_observed\_max)
    \end{split}
\end{align}
Although the above algorithm is carefully crafted and depends on the character of the predicates included in the step, it also contains a number of placeholders. Having filled those placeholders with predicate or operator translations from the domain-specific dictionary, it is possible to obtain strategy descriptions that accommodate for other problems. We treat Experiment~2 as a proof of concept that shows this for one class of problems, and adapt the dictionary such a way, as to have our method output instructions people understand. In more detail, the specific dictionary we used to transform this formula into task-dependent instructions was informed by pilot studies in which we tested multiple options and selected wording that resulted in the highest overall compliance. These natural language instructions are detailed in the next section, in which we assess the benefits of conveying the found strategies in behavioral experiments.

\section{Experiment 2: Boosting human performance in naturalistic decision-making and planning tasks with AI-generated decision aids}
Our previous work \citep{skirzynski2020automatic} showed that the static descriptions generated by AI-Interpret improve the performance of individuals in the Mouselab-MDP task. However, those improvements were lower in environments which required more complex planning strategies. This was partly due to the fact that static descriptions of complex strategies are more difficult to understand. Experiment~1, which in fact tested the interpretability of the most complex strategy of \cite{skirzynski2020automatic}, showed that procedural instructions are easier to understand than static descriptions. In addition, the results of Experiment~1 suggest that our updated decision aids come with additional benefits over static descriptions. First, it takes less time to follow procedural instructions than to use decision aids requiring you to evaluate each planning step individually. Second, in contrast to static descriptions, procedural instructions require no introduction on how to apply them.

Equipped with these improvements, we test in this section whether decision aids generated by our extended AI-powered boosting method can enhance human performance in tasks that are more naturalistic than those used by \citeauthor{skirzynski2020automatic} (\citeyear{skirzynski2020automatic}). Concretely, we evaluate our approach on two naturalistic tasks. In the Road Trip task \citep{Callaway2022Rational}, participants are asked to plan an inexpensive trip by looking up hotel prices across cities visited during the trip (see Fig.~\ref{fig:envs}A). In the Mortgage task, participants are asked to choose a mortgage based on the interest rates of the available options in the first year, the following five years, and the following 15 years, respectively (see Fig.~\ref{fig:envs}B). Similar to the Mouselab-MDP task with increasing variance, the rewards in the Road Trip task vary the most at the potential final destinations. This reward structure favors far-sighted planning. This task allows us to test if AI-powered boosting can help people become more far-sighted because previous work found that human planning in this task is more short-sighted than the optimal planning strategy \citep{Callaway2022Rational}. The same is also true of the Mortgage task. We designed this task so that the most long-term financial consequences of choosing a mortgage are the most crucial for the total cost. It can therefore be seen as a more naturalistic measure of people's shortsightedness in intertemporal choice \citep{o2015present,meier2010present}.

To test the benefits of using our extended AI-powered boosting approach, we conducted a large-scale online experiment in which participants were presented with the Road Trip task and the Mortgage task. To measure the benefits conferred by our AI-generated decision aids, we compare the performance of people being assisted by the automatically generated decision aids that conveyed a far-sighted strategy described in the previous section against the performance of a control group making decisions without a decision aid. 
We did not include a condition with static instructions because Experiment~1 conclusively showed that static instructions are less effective than procedural instructions. In Experiment~2, we also overcome the main limitation of Experiment~1 by studying how beneficial AI-powered boosting might be in real-world settings where people can choose to ignore the decision aid. Thus, instead of requiring participants to comply with the recommended strategy, Experiment~2 provides people with a decision aid and allows participants to freely decide how they want to make their decisions.

\subsection{Methods}

\subsubsection{Participants}
We recruited 111 people on Prolific. The mean duration of the experiment was 18.1 minutes in the control condition and 17.9 in the experimental condition. We excluded 2 participants (1.8\%; both in the experimental condition) who needed more than 3 quiz attempts in one of the two quizzes. This yielded 54 participants for the control condition (average age: 38.0 years, range: 19-74 years; 39 female) and 55 participants for the experimental condition (average age: 35.6 years, range: 18-70 years; 37 female).

\subsubsection{Tasks}\label{sec:methods_exp2}

Participants were engaged in two different planning tasks. The Road Trip task (see Fig.~\ref{fig:envs}a) asked people to plan a route from a starting location to one of multiple airport cities. Each city on the route requires the traveler to rent a hotel for the night. The participant's task was to efficiently find a route with low accommodation costs. To do so, the participant could look up the price of the cheapest hotel in a city for a \$10 fee by typing the city name into a search engine. Hotel prices were drawn from a uniform distribution over the values $\{\$30, \$35, \$40, \$45\}$ and airport hotel prices were drawn from a uniform distribution over the values $\{\$260, \$290, \$320, \$350, \$380\}$. In addition, one randomly selected airport city offered a price of only \$20. A route could be submitted after selecting a sequence of roads connecting the start city with an airport city.

The Mortgage task (see Fig.~\ref{fig:envs}b) asked people to choose the cheapest mortgage out of three options presented in a table. Each mortgage was defined by three different interest rates: the interest rate for the fist year (2022), the interest rate for the following five years (2023 – 2027) and the interest rate for the 25 years after that (2028 – 2052). Three different mortgages were presented: 1. A mortgage whose interest rates increased over time. The interest rates were drawn from normal distributions with means 0.5\%, 1.5\% and 2.5\% for the three time periods, respectively. 2. A mortgage whose interest rates stayed constant over time. The interest rates were drawn from a normal distribution with mean 1.5\% for all three time periods. 3. A mortgage whose interest rates decreased over time. The interest rates were drawn from normal distributions with means 2.5\%, 1.5\% and 0.5\% and for the three time periods, respectively. All distributions had a standard deviation of 0.44\%. The minimum value of an interest rate was set to 0. The mortgage with decreasing interest rates offered the lowest overall interest rate payments and thus represented the best choice. Participants could reveal up to three different interest rates for no fee by clicking on the corresponding table cell. At each point in the task, the participant could decide for one mortgage plan and proceed to the next trial.

\subsubsection{Outcome measures}
We quantified far-sightedness by examining which information participants gathered in what order. We defined far-sighted planning as gathering information about the most long-term consequences first (i.e., the prices in airport cities in the Road Trip task and the interest rate for the last 25 years in the Mortgage task). We therefore measured far-sightedness by the proportion of such far-sighted planning operations among the first $k$ planning operations, where $k$ denotes the number of pieces of information that are available about the most-longterm consequences. We call this measure the far-sightedness quotient (FSQ). For example, consider a Road Trip trial that includes two possible final destinations: Choosing these two final destinations in the first 2 clicks results in an FSQ of $1$. Choosing one final destination and one stopover in the first two clicks results in an FSQ of $0.5$. If a person performed fewer planning operations than there were far-sighted planning options, $k$ was reduced to the number of performed planning operations. The values are reported as percentages.

We utilized the click agreement metric from Experiment~1 (see Section~\ref{sec:exptwodv}) to capture how well participants applied the far-sighted planning strategy recommended by the AI-generated decision aid. Whether a performed planning operation is consistent with the far-sighted planning strategy depends on the strategy's stopping rule. The far-sighted planning strategy stops planning if it encounters the best possible long-term outcome, which is given by a hotel price of \$20 in the Road Trip task and by an interest of 0\% in the Mortgage task.

In addition, we quantified participants' planning success. In the Road Trip task was measured the score per trial which was defined as the sum of lookup fees and route costs subtracted from the initial budget of \$500. In the Mortgage task, we measured participants' performance by whether they selected the mortgage plan with decreasing interest rates because its total cost was always the lowest. 

\subsubsection{Procedure}
\begin{figure}
\begin{center}
\includegraphics[width=0.9\linewidth]{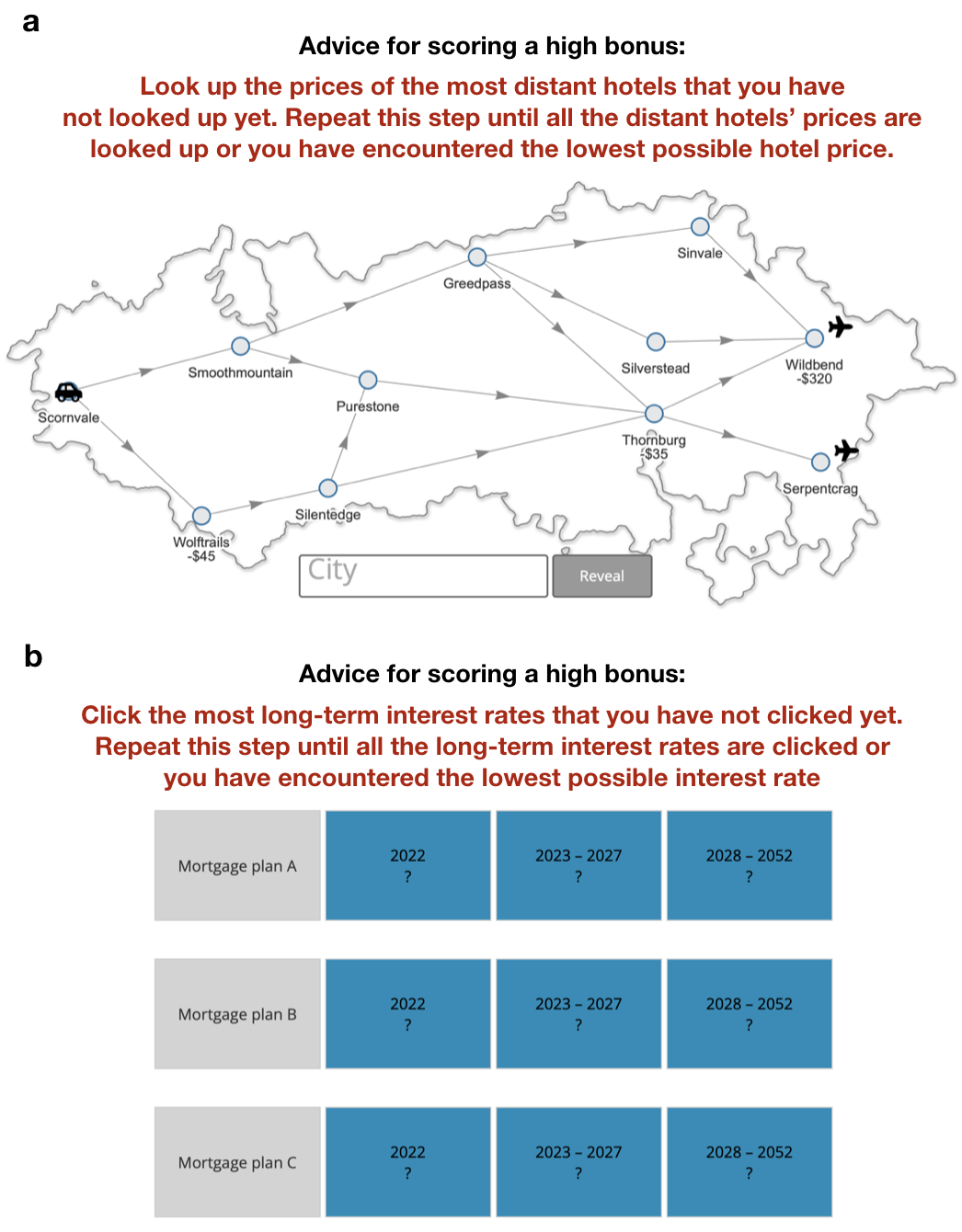}
\end{center}
\caption{Experiment 2: The Road Trip task (a) and the Mortgage task (b) with corresponding AI-generated decision aids as presented to the experimental condition. Only participants in the experimental condition were assisted with decision aids}
\label{fig:exp2_tasks}
\end{figure}

Participants were randomly divided into an experimental group and a control group. Each condition consisted of a Mortgage task block and a Road Trip task block, and a final demographic survey. The order in which the two task blocks were presented was randomized across participants. Both task blocks opened with instructions on the task (see Section \ref{appendix:exptwoTaskInstructions}), followed by a multiple-choice quiz on the task, an additional instructions page displayed only in the experimental condition (see Section \ref{appendix:exptwoTaskInstructions}), and eight trials of the actual task. The instructions of the Road Trip task explicitly stated that there is always a hotel in one of the airport cities with a rate of only \$20 per night. 
Participants received a bonus of 2 pence for every 100 points scored in the Road Trip task, or 5 pence per point scored in the Mortgage task. Additionally, everyone received a base payment of \pounds1.5. 
In the experimental condition, participants were assisted by our AI-generated decision aids. These decision aids were introduced as `Advice for scoring a high bonus' on the additional instructions page. Participants were told that the decision aids convey a near-optimal strategy for gathering information to arrive at a good decision. Participants were asked to try to understand the advice of the decision aid and how it can be applied to the task at hand. In addition, the AI-generated decision aids were displayed in the task trials at the top in red font with the note: `Advice to achieve a high bonus' (see. Fig.~\ref{fig:exp2_tasks}a-b). 

As described above, the AI-generated decision aids comprised procedural instructions for how to reach a decision. The procedural instructions for the Mortgage task were ``\emph{Click the most long-term interest rates that you have not clicked yet. Repeat this step until all the long-term interest rates are clicked or you have encountered the lowest possible interest rate.}'' The procedural instructions for the Road Trip task were ``\emph{Look up the prices of the most distant hotels that you have not looked up yet. Repeat this step until all the distant hotels' prices are looked up or you have encountered the lowest possible hotel price.}''

\subsubsection{Analysis}
We performed one-sided Mann-Whitney-U tests for group level comparisons according to our hypotheses that the experimental group would plan more far-sightedly and score higher than the control group. The study was pre-registered\footnote[1]{https://aspredicted.org/JRD_D7Z}. 

\subsection{Results}
\label{sec:results_exp2}
\begin{figure}[h!]
\begin{center}
\includegraphics[width=0.90\linewidth]{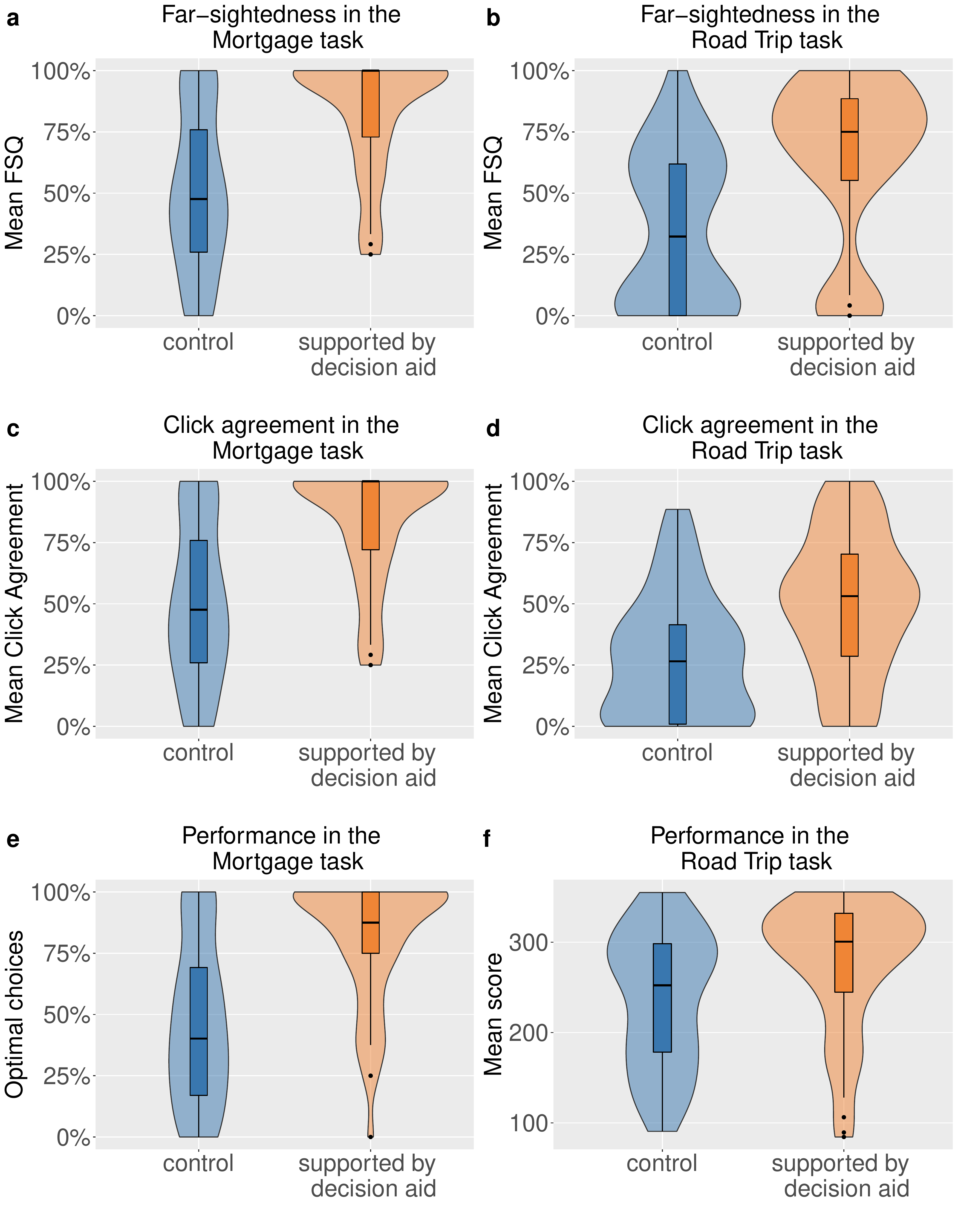}
\end{center}
 \vspace{-18pt}
\caption{Results of Experiment 2. The decision aids generated by our AI method boosted participants far-sightedness and performance in the Road Trip and Mortgage task. The experimental group was aided by the procedural instructions generated by our computational method. All differences are statistically significant (all $p \leq .01$).\\
Panels a-b: The experimental group planned more far-sighted than the control group. Panels c-d: The experimental group achieved higher click agreements than the control group. Panels e-f: Performance in both tasks was measured according to each task's objective. Higher scores mean better performance}
\label{fig:exp2_results}
\end{figure}

As illustrated in Fig.~\ref{fig:exp2_results}a-b, the experimental group planned significantly more far-sightedly than the control group in both tasks (Road Trip task: $A=.75$; \textit{U}=741, \textit{p}=<.001, Mortgage task: $A=.78$; \textit{U}=658, \textit{p}=<.001). In the Mortgage task, the automatically generated decision aid increased participants' average FSQ from 50.8\% in the control condition to 83.3\% in the experimental condition (medians: 47.6\% vs. 100\%). In the Road Trip task, the AI-generated decision aid increased the average FSQ from 35.1\% in the control condition to 63.8\% in the experimental condition (medians: 32.3\% vs. 75\%). In a follow-up analysis, we found that participants who had encountered the minimum airport price in one trial of the Road Trip task planned significantly more far-sightedly in the following trial compared to trials in which participants had not encountered the minimum airport price in the previous trial (FSQ: 74.7\% vs. 23.2\%; $A=.68$; \textit{U}=3203, \textit{p}=<.001).

The positive effect of AI-generated decision aids was also evident in participants' click agreement with the instructed strategy. Participants in the experimental group showed significantly higher click agreement than the control group in both tasks (Road Trip task: $A=.72$; \textit{U}=838, \textit{p}=<.001, Mortgage task: $A=.78$; \textit{U}=659, \textit{p}=<.001). In the Road Trip task, the average click agreement was 27.9\% in the control condition and 49.8\% in the experimental condition (medians: 26.5\% vs. 53.1\%; see Fig.~\ref{fig:exp2_results}d). In the Mortgage task, the stopping rule of the recommended strategy was never met; thus the mean click agreement only differed from the FSQ in the sense that it penalized not clicking all long-term options available. This was rarely the case; hence the click agreement was similar to the FSQ (control condition: 50.4\%; experimental condition: 82.9\% see Fig.~\ref{fig:exp2_results}c).

Furthermore, we found that the AI-generated decision aid significantly boosted participants' performance in both tasks (Road Trip task: $A=.66$; \textit{U}=994, \textit{p}=.001, Mortgage task: $A=.78$; \textit{U}=654, \textit{p}=<.001). On average a participant in the control condition selected the cheapest mortgage in only 46.8\% of trials (median: 40.2\%); by contrast participants in the experimental condition selected the cheapest mortgage 80.7\% of the time (median: 87.5\%; see Fig.~\ref{fig:exp2_results}e). In the Road Trip task, the AI-generated decision aid increased the participants' score from 237.4 points in the control condition (median: 252.3) to 276.3 points in the experimental condition (median: 300.6; see Fig.~\ref{fig:exp2_results}f).

Finally, we inspected how participants reached their decisions when they deviated from the recommended strategy. To do so, we averaged the frequency of the most common ways in which participants deviated from the optimal strategy across the two tasks. Out of every trial in which participants did not follow the recommended strategy, 21.1\% of the trials involved no clicks at all, in 35\% of trials participants started by inspecting an immediate outcome, in 12.5\% of trials participants started by clicking on an intermediate outcome, and in 31.4\% of trials participants started by inspecting a final outcome but then deviated from the optimal strategy in a later step. Two thirds of the time, the latter deviation already occurred in the second click.  

\section{Discussion and conclusion}
In this work, we developed a new approach to boosting human decision-making. While most previous work designed decision aids by hand, we extended a computational method that uses AI to generate decision aids automatically. Our method discovers and describes near-optimal strategies for human decision-making. The main technical contribution of this article was to develop an algorithm for transforming disjunctive normal form descriptions of planning strategies into procedural instructions for good decision-making that are easy to understand.

The results of Experiment 1 suggested that people can understand the kind of procedural instructions generated by our new method faster and follow the strategy more accurately than when it was conveyed by the decision aids generated by our previous method \citep{skirzynski2020automatic}. Moreover, we demonstrated that the decision aids generated by our new method can improve human decision-making in two naturalistic tasks: planning a road trip and choosing a mortgage. Our AI-generated decision aids improved the process and outcomes of people's decision-making in both tasks. This happened despite the fact people had the freedom to use their own decision strategy for those tasks. Presumably, people used the strategies hinted by our decision aid since they believed that it would be beneficial for them. To the best of our knowledge, this is the first demonstration that human decision-making in naturalistic tasks can be improved in this way by AI-generated natural language descriptions of near-optimal decision strategies. 

Together, the two major components of our AI-powered boosting method, that is AI-Interpret and DNF2LTL, give rise to a new algorithm for explainable reinforcement learning that can create descriptions of virtually any learned policy \citep{puiutta2020explainable, dazeley2021explainable}. Our experiments indicated that this new algorithm generates syntactically interpretable descriptions, that when translated into natural language, facilitate people's trust. People followed the policy and planned far-sightedly in Experiment~2 after only being presented with the output of AI-powered boosting as a decision aid they could use or ignore. Future work on explainable reinforcement learning should aim to test this new algorithm on a wider range of tasks and environments.

The goal of this article was to develop a method for improving human decision-making. We therefore evaluated our overall approach by whether and to which extent the resulting decision aids leads to better decisions. This metric depends on how much better the resource-rational strategy for a given decision problem is than people’s intuitive strategy, the method used to discover the resource-rational strategy, and how accurately the automatically generated natural language description captures the essence of the discovered strategy. Each of those components can be assessed separately. For an in-depth evaluation of how accurately the natural language instructions generated by our method describe the different decision strategies, please see \cite{skirzynski2021automatic}. 

Our experiments demonstrate that the decision aid boosted people's performance in the assisted decisions by guiding them through the process of executing a resource-rational decision strategy. This strategy ensured that participants considered the most important consequences of the actions they were choosing between. It thereby guided them to utilize crucial information that they might not have seen otherwise. In principle, this improvement might have occurred solely because the participant blindly followed the instructions of the decision aid. However, it is also conceivable that practicing good decision-making with a decision aid improves people's decision-making competency. Concretely, people's decision-making competence might improve because they internalize the described decision strategy through repetition or because they gain insights into the logic of the conveyed strategy and why it is adaptive.  If that were the case, then they might also use the conveyed strategy in future decisions they make without the decision aid. In that case, our decision aid would have boosted not just their performance in the task, but also their decision-making competence. Given that people can learn and transfer adaptive decision strategies through practice \citep{CognitiveTutorsPNAS,He2021measuring,He2022Where}, repeatedly using our decision aids might indeed have led to learning-induced improvements in their decision-making competency. Testing whether such learning occurs is an important direction for future research.

Whether our decision aid boosted people's decision-making competence is closely related to the question of why and how its provision improved people's decisions. There are at least two possible mechanisms: insight versus compliance. According to the first hypothesis (i.e., insight), the decision aid helped people gain insights into the logic of good decision-making that they then autonomously applied to improve their performance. Concretely, people might try out the recommended strategy because they have some level of trust in such recommendations. Because the recommended strategy is highly adaptive, the experienced outcomes will likely convince the participants that using the strategy is beneficial for them. Moreover, they might realize that the far-sighted strategy is adaptive because the final outcomes are more variable than earlier outcomes (insight). Based on that, they might then choose to continue using the strategy because they conclude that it works well and makes sense. According to the second hypothesis, participants interpreted the decision aid as a series of orders that they felt obliged to follow. Moreover, participants might have followed those instructions without understanding why they make any sense. We think that the most extreme version of this interpretation is unlikely because Experiment~2 emphasized the participants' autonomy by framing the decision aid as advice (see Figure~\ref{fig:exp2_tasks}). However, participants often comply with what they perceive to be the experimenter's implicit expectations, regardless of what the experimenter's actual intent is \citep{Orne1996demand}. Moreover, our experiments were not designed to distinguish between the two interpretations. Therefore, the mechanism through which the observed improvements were attained remains unclear. To address this important question, future studies should measure why participants followed the recommended strategy and test participants' understanding of the logic behind the recommended strategy and why it is effective. Based on existing measures of autonomous motivation \citep{sheldon2004self}, future studies could measure why participants followed the recommended strategy by asking them to rate to which extent they followed the strategy because they felt that it was required or expected of them and to separately rate to which extent they followed it because it made sense to them and because they thought it was a good strategy.

Since we designed our algorithm for generating natural language descriptions of decisions strategies with a particular class of problems in mind and only tested it on those problems, it remains unclear how well it would work for other kinds of decision problems. The principles are general enough that they are applicable to a wide range of decision problems. However, the dictionary is application-specific, and the implementation uses heuristics that probably will not work well for all possible applications. Therefore,  depending on how different future applications will be from the ones we tackled in this article, our method will require some amount of adaptation. Nevertheless, our work provides a proof of concept that it is possible to  discover and describe rational decision strategies automatically. Developing a more principled and more general translation algorithm is an important direction for future work.

Based on our positive findings on improving people's decision-making, we believe that future research should focus on even more realistic tasks and decision support in the real world. In this work, we found that the near-optimal decision strategy that the automatic strategy discovery method  \citep{CognitiveTutorsPNAS,skirzynski2020automatic} discovered for a simple Mouselab-MDP task (see Fig.~\ref{fig:mouselab}) could be automatically translated into effective decision aids for two more complex and more naturalistic tasks (see Fig.~\ref{fig:envs}). This worked not only for the Road Trip task that is structurally similar to the Mouselab-MDP task, but also for the problem of choosing a mortgage that is analogous to the Mouselab-MDP task at a more abstract level. This suggests that our method is, in principle, applicable to a wide range of decision-problems people face in the real-world as long as the essential structure of those problems can be modelled within our general metalevel MDP framework \citep{Griffiths2019,Callaway2022Rational,CognitiveTutorsPNAS}. We have previously argued that this is the case for a wide-range of common real-world decisions, such as purchasing decisions, hiring choices, investment decisions, deciding which charity to donate to, medical diagnosis, treatment planning, and credit approval decisions \citep{RSDArticle,skirzynski2020automatic,consul2021improving,CognitiveTutorsPNAS}. 

Applying our approach to such real-world problems requires building models of real-world decisions. Developing such models generally requires making evidence-based assumptions about the structure of the real world. Our knowledge of the real-world problems in which people have to make decisions are inevitably uncertain, sometimes inaccurate, and usually incomplete \citep{hertwig2019taming}. However, our approach does not require that all of those assumptions are correct. To the contrary, the methods described here can be combined with recent advances that have made strategy discovery methods robust to errors in the model of the decision problems to be solved \citep{RSDArticle}. 
Moreover, our strategy discovery methods can also be extended to environments where the true state of affairs is unknown because it cannot be observed directly \citep{heindrich2022PO}. In practice, the applicability of our method also depends on the size of the model. However, technical advances in machine learning methods for automatic strategy discovery are making our approach increasingly more scalable \citep{consul2021improving}. 
Regardless thereof, making AI-powered boosting work in the real world remains a difficult challenge. Whether future work will be able to overcome the remaining difficulties remains to be seen, but the results presented in this article make us cautiously optimistic.

\FloatBarrier

\section*{Declarations}

\paragraph{Funding}
This work was supported by the German Federal Ministry of Education and Research (BMBF): Tübingen AI Center, FKZ: 01IS18039B. This work was additionally supported by the Cyber Valley Research Fund (CyVy-RF-2019-02).

\paragraph{Conflicts of interest/Competing interests}
On behalf of all authors, the corresponding author states that there is no conflict of interest.

\paragraph{Availability of data and material (data transparency)}
The datasets analyzed for this study, alongside the code for the analysis can be found at \url{https://github.com/RationalityEnhancement/InterpretableStrategyDiscovery}.

\paragraph{Code availability (software application or custom code)}
The code for the algorithm introduced in the paper is available at \url{https://github.com/RationalityEnhancement/InterpretableStrategyDiscovery/tree/master/DNF2LTL}.

\paragraph{Ethics approval}
The experiments reported in this article were approved by the IEC of the University of T\"ubingen under IRB protocol number 667/2018BO2 (``Online-Experimente \"uber das Erlernen von Entscheidungsstrategien'').

\paragraph{Consent to participate (include appropriate statements)}
All participants gave their informed consent before starting the experiments.
\paragraph{Consent for publication}
All authors consent to the publication of this manuscript.

\bibliography{test}

\newpage

\section*{Appendix}

\renewcommand\thesection{A.\arabic{section}}    
\setcounter{section}{0}

\section{Experiment 1: Task instructions}
\label{appendix:exponeTaskInstructions}
Experiment 1 consisted of a condition that was assisted by static descriptions (flowchart) and a condition that was assisted by procedural instructions. To inform participants how to use the static descriptions in the task at hand, we used the following instructions:

``Flowchart: The description of the strategy takes the form of a flowchart. It walks you through a list of one or more questions that you need to answer to by looking at the tree, and describes which nodes to click. Look at the image below to see how a flowchart can look like. Task procedure: 1. Read the flowchart carefully. 2. Think of a node you would like to click 3. Go through the flowchart and answer questions about that node. 4. Click that node if the flowchart landed you in a 'Click it' decision. Otherwise, think of a different node. 5. Once you are sure that you clicked all the nodes the flowchart allows clicking -- that is it would evaluate to 'Don't click it' for every node -- click Next to advance to the next trial.''

To inform participants how to use the procedural instructions in the task at hand, we used the following instructions:

``Instructions: The description of the strategy is conveyed as a sequence of instructions. The instructions tell you what to click step by step. Look at the image below to see how a sequence of instructions can look like. Task procedure To enact the strategy conveyed by the instructions, do the following: 1. Read the instructions carefully. 2. Click the nodes by following the procedure described in the instructions. 3. Once there are no more nodes the instructions allow clicking, click Next to advance to the next trial.''

\section{Experiment 1: Development of click agreement over time}

\begin{figure}[h!]
\begin{center}
\includegraphics[width=1\linewidth]{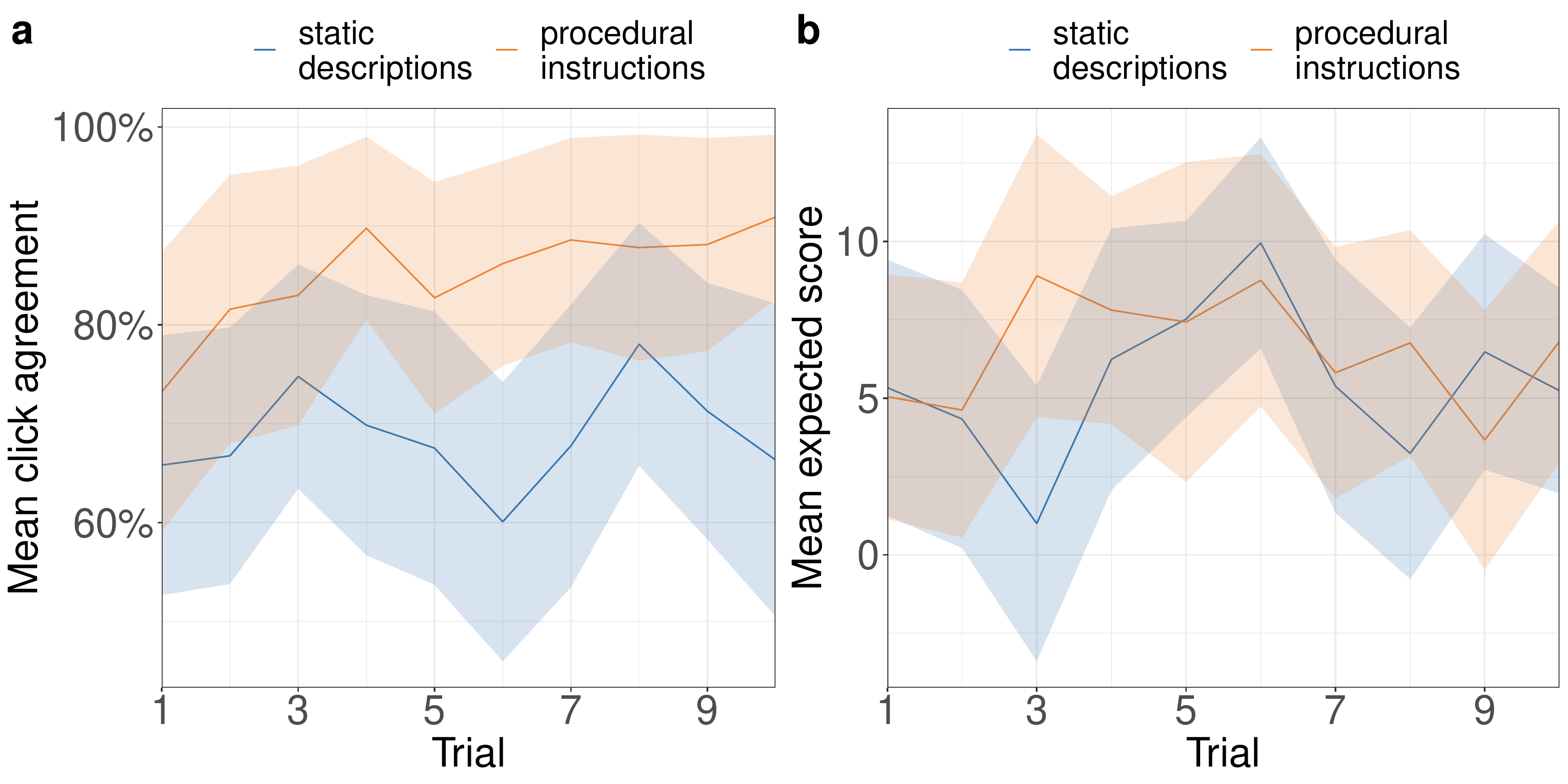}
\end{center}
\caption{Experiment 1: a) Procedural instructions are more interpretable than flowcharts. The plot shows the mean click agreement of participants per condition as a function of trial. Click agreement with the instructed clicking strategy was measured as the proportion of consistent clicks out of all performed clicks. b) The mean expected score that per condition as a function of trial. The shaded areas mark the 95\% confidence intervals
}
\label{fig:exp1_learningcurves}
\end{figure}

The development of click agreement over time is illustrated in Figure~\ref{fig:exp1_learningcurves}a. A linear regression model found that the click agreement in the static descriptions condition kept stable over time ($\beta = 0.002, p=.66$), whereas it significantly increased in the procedural instructions condition ($\beta=0.013, p= .002$), suggesting that the procedural instructions condition learned to apply the instructed strategy more precisely over time. Further, we found that the expected score (see Figure~\ref{fig:exp1_learningcurves}b) did not systematically change over time in the static descriptions condition ($\beta = 0.152, p=.487$) and in procedural descriptions condition ($\beta = -0.038, p=.87$). Lastly, we found that the fitted intercepts did not significantly differ between conditions (click agreement: $\beta = 0.1, p=.198$; expected score: $\beta = 2.1, p=.279$), indicating that there were no differences in the initial understanding of the task between conditions. 

\section{Experiment 2: Task instructions}
\label{appendix:exptwoTaskInstructions}

These were the instructions on the Road Trip Task: ``In the Travel Planner game, you pretend to be a travel planner. You start by seeing a map as shown below. Your client needs to travel from the city with the car (Ruby Ridge) to one of the cities with an airport. Getting from city A to city B is only possible when there is an arrow from city A to city B. Your client can travel only one city per day. During the night, he or she stays in a hotel, which costs money. Your client wants a morning flight, so they must pay for a hotel in the airport city as well. The price of the hotel varies between the different cities. Airport hotels start at \$20.Your client is on a tight budget of \$500 and wishes to take the cheapest route. Your goal is to choose which cities to traverse so that the price of the trip was as cheap as possible. You can look up the price of the cheapest hotel in a city by typing the city name in a text box and clicking Reveal. The prices are negative to convey the cost you will incur by staying in the city.When you look up a city, its price is revealed on the map. Revealing the price costs \$10. At any time, you can select parts of the client's route by clicking on the arrows. If you change your mind, you can unselect arrows by clicking them again. When you have finalized your route, click Submit. You do not need to check the prices of every city on the route before submitting.''

In addition, the experimental group received this information: ``Advice for scoring a high bonus: To help you score higher in the roadtrip planner game, we will show you its near-optimal strategy. This strategy describes in what order to explore the hotel prices. Please take a moment to understand this advice and how you could apply it in the game. Look up the prices of the most distant hotels that you have not looked up yet. Repeat this step until all the distant hotels' prices are looked up or you have encountered the lowest possible hotel price.''

These were the instructions on the Mortgage task: ``In the Mortgage game, you have found your dream property and want to ask the bank for a loan. The bank presents you with three different mortgage plans. Each mortgage plan has three different interest rates: One for the 1st year (2022), one for the 2nd until 5th year (2023 – 2027) and one for the 6th until 30th year (2028 – 2052). Unfortunately, the bank clerk forgot to tell you about the interest rates. In the example below, you can see three plans (Mortgage plan A, Mortgage plan B, Mortgage plan C) but their corresponding interest rates are hidden underneath the blue fields. You decide to call the bank to ask about the interest rates. However, the bank clerk only has time to tell you up to three interest rates. Each time a bank clerk tells you about the interest rate corresponds to one click. That means you can only click up to 3 times. Below you will see an example with one field revealed. In the example, the interest rate from 2023 to 2027 for mortgage plan  B was revealed. In the example, you would have to pay 1.61\% interest rate in each of the 4 years when you select mortgage plan B. You can click up to 3 times, after which you have to make a decision which mortgage plan to choose. You can select a mortgage plan at any time by clicking on the grey mortgage plan button (A, B or C).''

In addition, the experimental group received this information: ``Advice for scoring a high bonus: To help you score higher in the mortgage game, we will show you its near-optimal strategy. This strategy describes in what order to explore the interest rates. Please take a moment to understand this advice and how you could apply it in the game: Click the most long-term interest rates that you have not clicked yet. Repeat this step until all the long-term interest rates are clicked, or you have encountered the lowest possible interest rate.''

\section{Experiment 2: Development of FSQ over time}
\begin{figure}[h!]
\begin{center}
\includegraphics[width=1\linewidth]{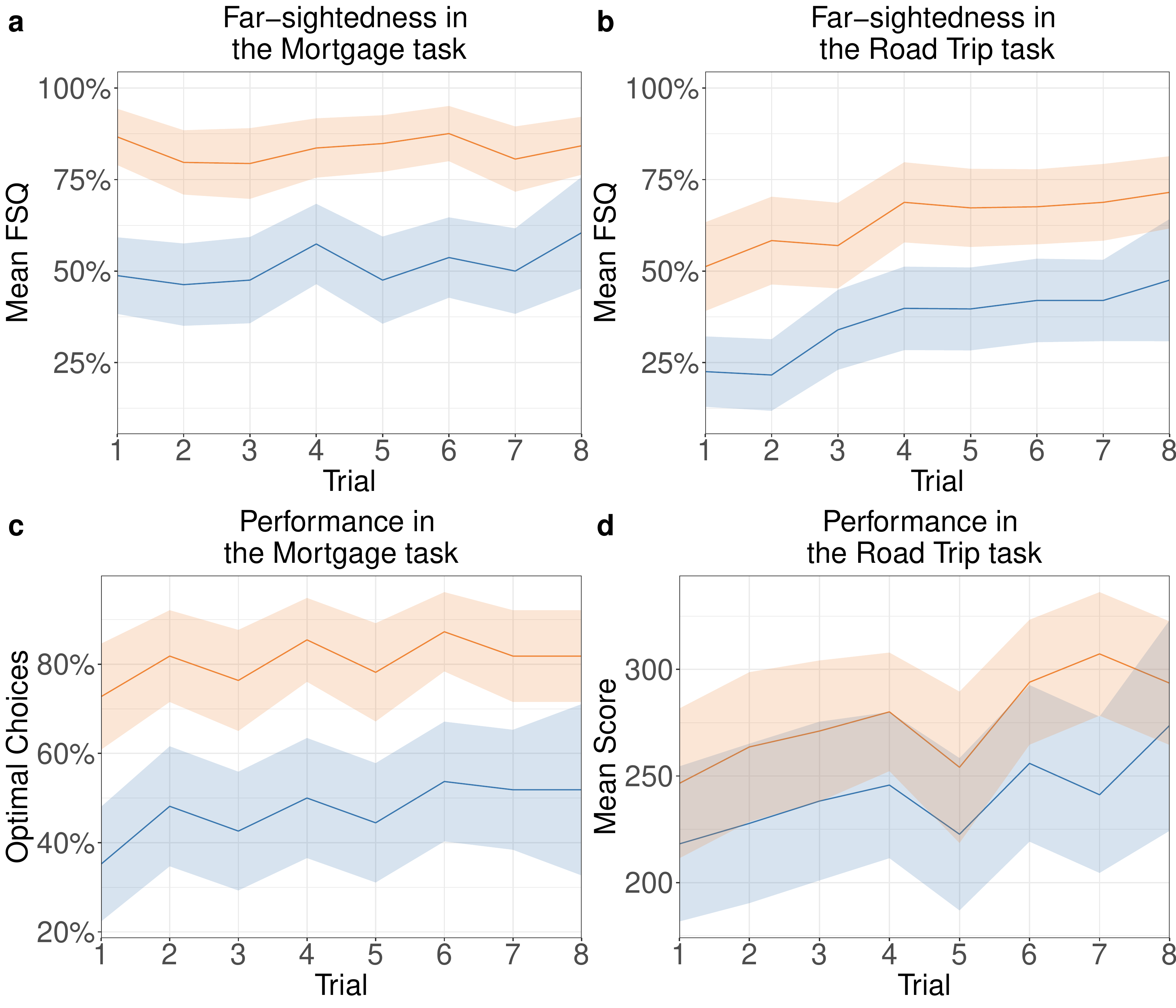}
\end{center}
\caption{Experiment 2: The experimental group (red) was supported by our decision aid whereas the control group (blue) was not. a) The plot shows the average far-sightedness quotient as a function of trial per condition for the Mortgage task. b) The plot shows the average far-sightedness quotient as a function of trial per condition for the Road Trip task. c) The plot shows the proportion of optimal choices as a function of trial per condition for the Mortgage task. b) The plot shows the average score as a function of trial per condition for the Mortgage task. The shaded areas mark the 95\% confidence intervals
}
\label{fig:exp2_learningcurves}
\end{figure}

In an exploratory analysis, we regressed the participant's FSQ in each task on the predictors \textit{decision aid} and \textit{trial number} and their interaction. We found that the intercept was significantly larger in the experimental condition than in the the control condition in both tasks (Mortgage task: $\beta = 0.35, p<.001$; Road Trip task: $\beta = 0.31, p<.001$), suggesting that the provision of the decision aid led to an immediate improvement in farsightedness. As illustrated in Figure~\ref{fig:exp2_learningcurves}, we found that in the Mortgage task the FSQ kept stable over time (trial number: $\beta = 0.01, p=.075$), whereas we found that the FSQ increased over time in the Road Trip task (trial number: $\beta = 0.03, p<.001$). The interaction of trial number and decision aid was insignificant for both tasks (Mortgage task: $\beta = 0.00, p=.266$; Road Trip task: $\beta = 0.00, p=.357$), indicating that the presence of the decision aid did not limit the learning. 

\section{Experiment 2: Development of performance over time}

In addition, we regressed the participants' performance in each task on the predictors \textit{decision aid} and \textit{trial number} and their interaction. We found that the intercept was significantly larger in the decision aid condition than in the no aid condition in the Mortgage task, but not in the Road Trip task (Mortgage task: $\beta = 2.5, p<.001$; Road Trip task: $\beta = 29.4, p=.1$). As illustrated in Figure~\ref{fig:exp2_learningcurves}c, we found that in the Mortgage task, the number of optimal choices increased over time (trial number: $\beta = 0.12, p=.034$). As illustrated in Figure~\ref{fig:exp2_learningcurves}d, we found that in the Road Trip task the score increased over time, however not significantly (trial number: $\beta = 4.7, p=.068$). The interaction of trial number and decision aid was insignificant for both tasks (Mortgage task: $\beta = -0.02, p=.785$; Road Trip task: $\beta = 2.3, p=.509$).

\end{document}